# Gap Safe Screening Rules for Sparsity Enforcing Penalties

**Eugene Ndiaye**                    EUGENE.NDIAYE@TELECOM-PARISTECH.FR
*LTCI, Télécom ParisTech, Université Paris-Saclay, 75013, Paris, France*

**Olivier Fercoq**                    OLIVIER.FERCOQ@TELECOM-PARISTECH.FR
*LTCI, Télécom ParisTech, Université Paris-Saclay, 75013, Paris, France*

**Alexandre Gramfort**                    ALEXANDRE.GRAMFORT@INRIA.FR
*Inria, Université Paris-Saclay, 91120, Palaiseau, France*
*LTCI, Télécom ParisTech, Université Paris-Saclay, 75013, Paris, France*

**Joseph Salmon**                    JOSEPH.SALMON@TELECOM-PARISTECH.FR
*LTCI, Télécom ParisTech, Université Paris-Saclay, 75013, Paris, France*

## Abstract

In high dimensional regression settings, sparsity enforcing penalties have proved useful to regularize the data-fitting term. A recently introduced technique called *screening rules* propose to ignore some variables in the optimization leveraging the expected sparsity of the solutions and consequently leading to faster solvers. When the procedure is guaranteed not to discard variables wrongly the rules are said to be *safe*. In this work, we propose a unifying framework for generalized linear models regularized with standard sparsity enforcing penalties such as $\ell_1$ or $\ell_1/\ell_2$ norms. Our technique allows to discard safely more variables than previously considered safe rules, particularly for low regularization parameters. Our proposed Gap Safe rules (so called because they rely on duality gap computation) can cope with any iterative solver but are particularly well suited to (block) coordinate descent methods. Applied to many standard learning tasks, Lasso, Sparse-Group Lasso, multi-task Lasso, binary and multinomial logistic regression, etc., we report significant speed-ups compared to previously proposed safe rules on all tested data sets.

## 1. Introduction

The computational burden of solving high dimensional regularized regression problem has led to a vast literature on improving algorithmic solvers in the last two decades. With the increasing popularity of $\ell_1$-type regularization ranging from the Lasso (Tibshirani, 1996) or group-Lasso (Yuan and Lin, 2006) to regularized logistic regression and multi-task learning, many algorithmic methods have emerged to solve the associated optimization problems (Koh et al., 2007; Bach et al., 2012). Although for the simple $\ell_1$ regularized least square a specific algorithm (*e.g.*, the LARS (Efron et al., 2004)) can be considered, for more general formulations, penalties, and possibly larger dimensions, (block) coordinate descent has proved to be an efficient strategy (Friedman et al., 2010).

Our main objective in this work is to propose a technique that can speed-up any iterative solver for such learning problems, and that is particularly well suited for (block) coordinate descent method as this type of method can easily ignore useless coordinates[1].

The *safe rules* introduced by El Ghaoui et al. (2012) for generalized $\ell_1$ regularized problems, is a set of rules allowing to eliminate features whose associated coefficients are guaranteed to be zero at the optimum, even before starting any algorithm. Relaxing the safe rule, one can obtain some additional speed-up at the price of possible mistakes. Such heuristic strategies, called *strong rules* by Tibshirani et al. (2012) reduce the computational cost using an active set strategy, but require multiple post-processing to check for features possibly wrongly discarded.

Another road to speed-up screening method has been pursued following the introduction of *sequential safe rules* (El Ghaoui et al., 2012; Wang et al., 2012; Xiang et al., 2014; Wang et al., 2014). The idea is to improve the screening thanks to the computation done for a previous regularization parameter as in homotopy/continuation methods. This scenario is particularly relevant in machine learning, where one computes solutions over a grid of regularization parameters, so as to select the best one, *e.g.,* by cross-validation. Nevertheless, the aforementioned methods suffer from the same problem as strong rules, since relevant features can be wrongly disregarded. Indeed, sequential rules usually rely on the exact knowledge of certain theoretical quantities that are only known approximately. Especially, for such rules to work one needs the exact dual optimal solution from the previous regularization parameter, a quantity (almost) never available to the practitioner.

The recent introduction of *dynamic safe rules* by Bonnefoy et al. (2015, 2014) has opened a new promising venue by performing variable screening, not only before the algorithm starts, but also along the iterations. This screening strategy can be applied for any standard optimization algorithm such as FISTA (Beck and Teboulle, 2009), primal-dual (Chambolle and Pock, 2011), augmented Lagrangian (Boyd et al., 2011). Yet, it is particularly relevant for strategies that can benefit from support reduction or active sets (Kowalski et al., 2011; Johnson and Guestrin, 2015), such as coordinate-descent (Fu, 1998; Friedman et al., 2007, 2010).

This paper contains a synthesis and a unified presentation of the methods introduced first for the Lasso in (Fercoq et al., 2015) and then for $\ell_1/\ell_2$ norms in (Ndiaye et al., 2015) as well as for Sparse-Group Lasso in (Ndiaye et al., 2016b). Our so-called Gap Safe rules (because the screening rules rely on duality gap computations), improved on dynamic safe rules for a broad class of learning problems with the following benefits:

- Gap Safe rules are easy to insert in existing solvers,

- they are proved to be safe and unify sequential and dynamic rules,

- they lead to improved speed-ups in practice *w.r.t.* previously known safe rules,

- they achieve fast variable identifications[2].

---

1. By construction a method like LARS (Efron et al., 2004), that applies only to the Lasso case, cannot beneficiate from screening rules.

2. more precisely, we identify faster the equicorrelation set, see Theorem 14

Furthermore, it is worth noting that strategies also leveraging dual gap computations have recently been considered to safely discard irrelevant coordinates: Shibagaki et al. (2016) have considered screening rules for learning tasks with both feature sparsity and sample sparsity, such as for $\ell_1$-regularized SVM. In this case, some interesting developments have been proposed, namely *safe keeping* strategies, which allow to identify features and samples that are guaranteed to be active. Constrained convex problems such as minimum enclosing ball can also be included as shown in Raj et al. (2016). The Blitz algorithm by Johnson and Guestrin (2015) aims to speed up working set methods using duality gaps computations; significant gains were also obtained in limited-memory and distributed settings.

We introduce the general framework of Gap Safe screening rules in Section 3; in Section 4 we instantiate them on various strategies including static, dynamic and sequential ones, and show how the Gap Safe methodology can encompass all of them. The converging nature of the Gap Safe rules is also discussed. In Section 5, we investigate the specific form of our rules for standard machine learning models: Lasso, Group Lasso, Sparse-Group Lasso, logistic regression with $\ell_1$ regularization, etc. Section 6 reports a comprehensive set of experiments on four different learning problems using either dense or sparse data. Results demonstrate the systematic gain in computation time of the Gap Safe rules.

## 2. Notation and Background on Optimization

For any integer $d \in \mathbb{N}$, we denote by $[d]$ the set $\{1, \ldots, d\}$ and by $Q^\top$ the transpose of a matrix $Q$. Our observation vector is $y \in \mathbb{R}^n$ where $n$ is the number of samples. The design matrix $X = [x_1, \ldots, x_n]^\top \in \mathbb{R}^{n \times p}$ has $p$ explanatory variables (or features) column-wise, and $n$ observations row-wise. For a norm $\Omega$, we write $\mathcal{B}_\Omega$ the associated unit ball, $\|\cdot\|_2$ is the $\ell_2$ norm (and $\langle \cdot, \cdot \rangle$ for the associated inner product), $\|\cdot\|_1$ is the $\ell_1$ norm, and $\|\cdot\|_\infty$ is the $\ell_\infty$ norm. The $\ell_2$ unit ball is denoted by $\mathcal{B}_2$ (or simply $\mathcal{B}$) and we write $\mathcal{B}(\theta, r)$ the $\ell_2$ ball with center $\theta$ and radius $r$. For a vector $\beta \in \mathbb{R}^p$, we denote by $\mathrm{supp}(\beta)$ the support of $\beta$ (*i.e.*, the set of indices corresponding to non-zero coefficients) and by $\|\mathrm{B}\|_F^2 = \sum_{j=1}^p \sum_{k=1}^q \mathrm{B}_{j,k}^2$ the Frobenius norm of a matrix $\mathrm{B} \in \mathbb{R}^{p \times q}$. We denote $(t)_+ = \max(0, t)$ and $\Pi_{\mathcal{C}}(\cdot)$ the projection operator over a closed convex set $\mathcal{C}$. The soft-thresholding operator $\mathrm{ST}_\tau$ (at level $\tau \geq 0$) is defined for any $x \in \mathbb{R}^d$ by $[\mathrm{ST}_\tau(x)]_j = \mathrm{sign}(x_j)(|x_j| - \tau)_+$.

The parameter to recover is a vector $\beta = (\beta_1, \ldots, \beta_p)^\top$ admitting a group structure. A group of features is a subset $g \subset [p]$ and $n_g$ is its cardinality. The set of groups is denoted by $\mathcal{G}$ and we focus only on non-overlapping groups[3] that form a partition of the set $[p]$. We denote by $\beta_g$ the vector in $\mathbb{R}^{n_g}$ which is the restriction of $\beta$ to the indices in $g$. We write $[\beta_g]_j$ the $j$-th coordinate of $\beta_g$. We also use the notation $X_g \in \mathbb{R}^{n \times n_g}$ to refer to the sub-matrix of $X$ assembled from the columns with indices $j \in g$ and $X_j$ when the groups are a single feature, *i.e.*, when $g = \{j\}$.

Some elements of convex analysis used in the following sections are introduced here. For a function $f : \mathbb{R}^d \to [-\infty, +\infty]$, the Fenchel-Legendre transform[4] of $f$, is the function $f^* : \mathbb{R}^d \to [-\infty, +\infty]$ defined by $f^*(u) = \sup_{z \in \mathbb{R}^d} \langle z, u \rangle - f(z)$. The sub-differential of a function $f$ at a point $x$ is denoted by $\partial f(x)$. For a norm $\Omega$ over $\mathbb{R}^d$, its dual norm is written $\Omega^D$ and is defined for any $u \in \mathbb{R}^d$ by $\Omega^D(u) = \max_{\Omega(z) \leq 1} \langle z, u \rangle$. Note that in the case of a group-decomposable norm, one can check that $\Omega^D(\beta) = \max_{g \in \mathcal{G}} \Omega_g^D(\beta_g)$ and $\partial\Omega(\beta) = \Pi_{g \in \mathcal{G}} \partial\Omega_g(\beta_g)$.

---

3. Overlapping groups could be treated as well, some insight is given in Section 5.3
4. Such a transform is often referred to as the (convex) conjugate of a (convex) function

We remind below useful standard results from convex analysis:

**Proposition 1 (Fermat's Rule)** *(see (Bauschke and Combettes, 2011, Proposition 26.1) for a more general result) For any convex function $f : \mathbb{R}^d \to \mathbb{R}$:*

$$x^\star \in \underset{x \in \mathbb{R}^d}{\arg\min}\, f(x) \Longleftrightarrow 0 \in \partial f(x^\star). \tag{1}$$

**Proposition 2 (Subdifferential of a Norm)** *(see (Bach et al., 2012, Proposition 1.2)) The sub-differential of the norm $\Omega$ at $x$, is given by*

$$\partial\Omega(x) = \begin{cases} \{z \in \mathbb{R}^d : \Omega^D(z) \leqslant 1\} = \mathcal{B}_{\Omega^D}, & \text{if } x = 0, \\ \{z \in \mathbb{R}^d : \Omega^D(z) = 1 \text{ and } z^\top x = \Omega(x)\}, & \text{otherwise.} \end{cases} \tag{2}$$

## 3. Gap Safe Framework

We propose to estimate the vector of parameters $\beta$ by solving

$$\hat{\beta}^{(\lambda)} \in \underset{\beta \in \mathbb{R}^p}{\arg\min}\, P_\lambda(\beta), \text{ for } P_\lambda(\beta) := F(\beta) + \lambda\Omega(\beta) := \sum_{i=1}^n f_i(x_i^\top \beta) + \lambda\Omega(\beta) \ , \tag{3}$$

where all $f_i : \mathbb{R} \mapsto \mathbb{R}$ are convex and differentiable functions with $1/\gamma$-Lipschitz gradient and $\Omega : \mathbb{R}^p \mapsto \mathbb{R}_+$ is a norm that is group-decomposable, *i.e.*, $\Omega(\beta) = \sum_{g \in \mathcal{G}} \Omega_g(\beta_g)$ where each $\Omega_g$ is a norm on $\mathbb{R}^{n_g}$. The $\lambda$ parameter is a non-negative constant controlling the trade-off between the data fitting term and the regularization term. Popular instantiations of problems of the form (3) are detailed in Section 5.

**Theorem 3** *A dual formulation of the optimization problem defined in (3) is given by*

$$\hat{\theta}^{(\lambda)} = \underset{\theta \in \Delta_X}{\arg\max} - \sum_{i=1}^n f_i^*(-\lambda\theta_i) =: D_\lambda(\theta), \tag{4}$$

*where $\Delta_X = \{\theta \in \mathbb{R}^n : \forall g \in \mathcal{G}, \Omega_g^D(X_g^\top \theta) \leqslant 1\}$. Moreover, the Fermat's rule reads:*

$$\forall i \in [n],\ \hat{\theta}_i^{(\lambda)} = -\nabla f_i(x_i^\top \hat{\beta}^{(\lambda)})/\lambda \quad \textbf{\textit{(link equation)}}, \tag{5}$$

$$\forall g \in \mathcal{G},\ X_g^\top \hat{\theta}^{(\lambda)} \in \partial\Omega_g(\hat{\beta}_g^{(\lambda)}) \quad \textbf{\textit{(sub-differential inclusion)}}. \tag{6}$$

For any $\theta \in \mathbb{R}^n$ let us introduce $G(\theta) := [\nabla f_1(\theta_1), \dots, \nabla f_n(\theta_n)]^\top \in \mathbb{R}^n$. Then the primal/dual link equation can be written $\hat{\theta}^{(\lambda)} = -G(X\hat{\beta}^{(\lambda)})/\lambda$.

Contrarily to the primal, the dual problem has a unique solution under our assumption on the $f_i$'s. Indeed, the dual function is strongly concave, hence strictly concave.

### 3.1 Safe Screening Rules

Following the seminal work by El Ghaoui et al. (2012) screening techniques have emerged as a way to exploit the known sparsity of the solution by discarding features prior to starting a sparse solver. Such techniques are referred to in the literature as *safe rules* when they screen out coefficients guaranteed to be zero in the targeted optimal solution. Zeroing those coefficients allows to focus exclusively on the non-zero ones (likely to represent signal) and helps reducing the computational burden.

One well known extreme is the following: for $\lambda > 0$ large enough, 0 is the unique solution of (3). Indeed,

$$0 \in \underset{\beta \in \mathbb{R}^p}{\arg \min} \, F(\beta) + \lambda \Omega(\beta) \overset{(1)}{\iff} 0 \in \{\nabla F(0)\} + \lambda \partial \Omega(0) \overset{(2)}{\iff} \Omega^D(\nabla F(0)) \leqslant \lambda .$$

Hence we recall the first "naive" screening rule, stating that for large values of the regularization parameter, all features can be discarded.

**Proposition 4 (Critical Parameter: $\lambda_{\max}$)** *For any $\lambda > 0$,*

$$0 \in \underset{\beta \in \mathbb{R}^p}{\arg \min} \, P_\lambda(\beta) \iff \lambda \geqslant \lambda_{\max} := \Omega^D(\nabla F(0)) = \Omega^D(X^\top G(0)) \ .$$

So from now on, we will only focus on the case where $\lambda < \lambda_{\max}$. In this case, screening rules rely on a direct consequence of Fermat's rule (6). If $\hat{\beta}_g^{(\lambda)} \neq 0$, then $\Omega_g^D(X_g^\top \hat{\theta}^{(\lambda)}) = 1$ thanks to Equation (2). Since $\hat{\theta}^{(\lambda)} \in \Delta_X$, it implies, by contrapositive, that if $\Omega_g^D(X_g^\top \hat{\theta}^{(\lambda)}) < 1$ then $\hat{\beta}_g^{(\lambda)} = 0$. This relation means that the $g$-th group can be discarded whenever $\Omega_g^D(X_g^\top \hat{\theta}^{(\lambda)}) < 1$. However, since $\hat{\theta}^{(\lambda)}$ is **unknown** — unless $\lambda > \lambda_{\max}$, in which case $\hat{\theta}^{(\lambda)} = G(0)/\lambda$ — this rule is of limited use. Fortunately, it is often possible to construct a set $\mathcal{R} \subset \mathbb{R}^n$, called a *safe region*, that contains $\hat{\theta}^{(\lambda)}$. This observation leads to the following result.

**Proposition 5 (Safe Screening Rule El Ghaoui et al. (2012))** *If $\hat{\theta}^{(\lambda)} \in \mathcal{R}$, and $g \in \mathcal{G}$:*

$$\max_{\theta \in \mathcal{R}} \Omega_g^D(X_g^\top \theta) < 1 \implies \Omega_g^D(X_g^\top \hat{\theta}^{(\lambda)}) < 1 \implies \hat{\beta}_g^{(\lambda)} = 0 \ . \tag{7}$$

The so-called *safe screening* rule consists in removing the $g$-th group from the problem whenever the previous test is satisfied, since then $\hat{\beta}_g^{(\lambda)}$ is guaranteed to be zero. Should $\mathcal{R}$ be small enough to screen many groups, one can observe considerable speed-ups in practice as long as the testing can be performed efficiently. A natural goal is to find safe regions as narrow as possible: smaller safe regions can only increase the number of screened out variables. To have useful screening procedures one needs:

- the safe region $\mathcal{R}$ to be as small as possible (and to contain $\hat{\theta}^{(\lambda)}$),

- the computation of the quantity $\max_{\theta \in \mathcal{R}} \Omega_g^D(X_g^\top \theta)$ to be cheap.

The later means that safe regions should be simple geometric objects, since otherwise, evaluating the test could lead to a computational burden limiting the benefits of screening.

### 3.2 Gap Safe Regions

Various shapes have been considered in practice for the safe region $\mathcal{R}$ such as balls (El Ghaoui et al., 2012), domes (Fercoq et al., 2015) or more refined sets (see Xiang et al. (2014) for a survey). Here we consider for simplicity the so-called "sphere regions" (following the terminology introduced by El Ghaoui et al. (2012)) choosing a ball $\mathcal{R} = \mathcal{B}(\theta, r)$ as a safe region. Thanks to the triangle inequality, we have:

$$\max_{\theta \in \mathcal{B}(\theta, r)} \Omega_g^D(X_g^\top \theta) \leqslant \Omega_g^D(X_g^\top \theta) + \max_{\theta \in \mathcal{B}(\theta, r)} \Omega_g^D(X_g^\top(\theta - \theta)),$$

and denoting $\Omega_g^D(X_g) := \sup_{u \neq 0} \frac{\Omega_g^D(X_g^\top u)}{\|u\|_2}$ the operator norm of $X_g$ associated to $\Omega_g^D(\cdot)$, we deduce from Proposition 7 the screening rule for the $g$-th group:

$$\text{Safe sphere test:} \qquad \text{If} \quad \Omega_g^D(X_g^\top \theta) + r\Omega_g^D(X_g) < 1, \quad \text{then} \quad \hat{\beta}_g^{(\lambda)} = 0 \ . \tag{8}$$

#### 3.2.1 FINDING A CENTER

To create a useful center for a safe sphere, one needs to be able to create dual feasible points, *i.e.*, points in the dual feasible set $\Delta_X$. One such point is $\theta_{\max} := -G(0)/\lambda_{\max}$ which leads to the original static safe rules proposed by El Ghaoui et al. (2012). Yet, it has a limited interest, being helpful only for a small range of (large) regularization parameters $\lambda$, as discussed in Section 4.1. A more generic way of creating a dual point that will be key for creating our safe rules is to rescale any point $z \in \mathbb{R}^n$ such that it is in the dual set $\Delta_X$. The rescaled point is denoted by $\Theta(z)$ and is defined by

$$\Theta(z) := \begin{cases} z, & \text{if } \Omega^D(X^\top z) \leqslant 1, \\ \frac{z}{\Omega^D(X^\top z)}, & \text{otherwise.} \end{cases} \tag{9}$$

This choice guarantees that $\forall z \in \mathbb{R}^n, \Theta(z) \in \Delta_X$. A candidate often considered for computing a dual point is the (generalized) residual term $z = -G(X\beta)/\lambda$. This choice is motivated by the primal-dual link equation 5 *i.e.*, $\hat{\theta}^{(\lambda)} = -G(X\hat{\beta}^{(\lambda)})/\lambda$.

#### 3.2.2 FINDING A RADIUS

Now that we have seen how to create a center candidate for the sphere, we need to find a proper radius, that would allow the associated sphere to be safe. The following theorem proposes a way to obtain a radius using the duality gap. The quantity

$$\text{Gap}_\lambda(\beta, \theta) := P_\lambda(\beta) - D_\lambda(\theta) \tag{10}$$

is often referred to as the duality gap in the convex optimization literature, hence the name of our proposed Gap Safe framework. This quantity is also a useful tool when designing a stopping criterion: noting that for any $\beta \in \mathbb{R}^p, \theta \in \Delta_X, P_\lambda(\beta) - P_\lambda(\hat{\beta}^{(\lambda)}) \leqslant \text{Gap}_\lambda(\beta, \theta)$, it suffices to find a primal-dual pair with a duality gap smaller than $\epsilon$ to ensure an $\epsilon$-accuracy primal solution for Problem 3.

**Theorem 6 (Gap Safe Sphere)** *Assuming that $F$ has $1/\gamma$-Lipschitz gradient, we have*

$$\forall \beta \in \mathbb{R}^p, \forall \theta \in \Delta_X, \quad \left\| \hat{\theta}^{(\lambda)} - \theta \right\|_2 \leqslant \sqrt{\frac{2\text{Gap}_\lambda(\beta, \theta)}{\gamma \lambda^2}} \ . \tag{11}$$

*Hence the set $\mathcal{R} = \mathcal{B}(\theta, \sqrt{2\mathrm{Gap}_\lambda(\beta, \theta)/\gamma\lambda^2})$ is a safe region for any $\beta \in \mathbb{R}^n$ and $\theta \in \Delta_X$.*

**Proof** Remember that $\forall i \in [n], f_i$ is differentiable with a $1/\gamma$-Lipschitz gradient. As a consequence, $\forall i \in [n], f_i^*$ is $\gamma$-strongly convex (Hiriart-Urruty and Lemaréchal, 1993, Theorem 4.2.2, p. 83) and so the dual function $D_\lambda$ is $\gamma\lambda^2$-strongly concave:

$$\forall (\theta_1, \theta_2) \in \mathbb{R}^n \times \mathbb{R}^n, \quad D_\lambda(\theta_2) \leqslant D_\lambda(\theta_1) + \langle \nabla D_\lambda(\theta_1), \theta_2 - \theta_1 \rangle - \frac{\gamma\lambda^2}{2} \|\theta_1 - \theta_2\|_2^2 \ .$$

Specifying the previous inequality for $\theta_1 = \hat{\theta}^{(\lambda)}, \theta_2 = \theta \in \Delta_X$, one has

$$D_\lambda(\theta) \leqslant D_\lambda(\hat{\theta}^{(\lambda)}) + \langle \nabla D_\lambda(\hat{\theta}^{(\lambda)}), \theta - \hat{\theta}^{(\lambda)} \rangle - \frac{\gamma\lambda^2}{2} \left\| \hat{\theta}^{(\lambda)} - \theta \right\|_2^2 \ .$$

By definition, $\hat{\theta}^{(\lambda)}$ maximizes $D_\lambda$ on $\Delta_X$, so, $\langle \nabla D_\lambda(\hat{\theta}^{(\lambda)}), \theta - \hat{\theta}^{(\lambda)} \rangle \leqslant 0$. This implies

$$D_\lambda(\theta) \leqslant D_\lambda(\hat{\theta}^{(\lambda)}) - \frac{\gamma\lambda^2}{2} \left\| \hat{\theta}^{(\lambda)} - \theta \right\|_2^2.$$

By weak duality $\forall \beta \in \mathbb{R}^p, D_\lambda(\hat{\theta}^{(\lambda)}) \leqslant P_\lambda(\beta)$, hence $\forall \beta \in \mathbb{R}^p, \forall \theta \in \Delta_X, D_\lambda(\theta) \leqslant P_\lambda(\beta) - \frac{\gamma\lambda^2}{2}\|\hat{\theta}^{(\lambda)} - \theta\|_2^2$ and the conclusion follows. ∎

**Remark 7** *To build a Gap Safe region as in Equation (11), we only need strong convexity in the dual which is equivalent to smoothness of the loss function whereas the screening property (7), requires group separability of norms. Hence our framework of Gap Safe screening rule automatically applies for a large class of problems.*

**Remark 8** *During the review process, we became aware of a possible improvement for the radius Johnson and Guestrin (2016). In the Blitz framework, their approach leads to a potentially smaller radius, using a strongly concave upper bound of the dual function whose maximum is known. In our framework, this can be used to improve the safe radius by a $\sqrt{2}$ factor in the static case. This is unclear to us whether this can be done for the sequential and dynamic version. For the SVM problem Zimmert et al. (2015) got the same improvement by writing the duality gap as a function of primal variables only.*

### 3.2.3 Safe Active Set

Note that any time a safe rule is performed thanks to a safe region $\mathcal{R} = \mathcal{B}(\theta, r)$, one can associate a *safe active set* $\mathcal{A}_{\theta, r}$, consisting of the features that cannot be removed yet by the test in Equation (8). Hence, the safe active set contains the true support of $\hat{\beta}^{(\lambda)}$.

**Definition 9 (Safe Active Set)** *For a center $\theta \in \Delta_X$ and a radius $r \geqslant 0$, the safe (sphere) active set consists of the variables not eliminated by the associated (sphere) safe rule,* i.e.,

$$\mathcal{A}(\theta, r) := \{g \in \mathcal{G} : \Omega_g^D(X_g^\top \theta) + r\Omega_g^D(X_g) \geqslant 1\} \ . \tag{12}$$

When choosing $z = -G(X\beta)/\lambda$ as proposed in Section 3.2.1 as the current residual, the computation of $\theta = \Theta(z)$ in Equation (9) involves the computation of $\Omega^D(X^\top z)$. A straightforward implementation would cost $\mathcal{O}(np)$ operations. This can be avoided: when using a safe rule one knows that the index achieving the maximum for this norm is in $\mathcal{A}(\theta, r)$. Indeed, by construction of the safe active set, it is easy to see that $\Omega^D(X^\top z) = \max_{g \in \mathcal{A}(\theta, r)} \Omega_g^D(X_g^\top z)$. In practice the evaluation of the dual gap is therefore $\mathcal{O}(nq)$ where $q$ is the size of $\mathcal{A}(\theta, r)$. In other words, using a safe screening rule also speeds up the evaluation of the stopping criterion.

### 3.3 Outline of the Algorithm

When designing a supervised learning algorithm with sparsity enforcing penalties, the tuning of the parameter $\lambda$ in Problem (3) is crucial and is usually done by cross-validation which requires evaluation over a grid of parameter values. A standard grid considered in the literature is $\lambda_t = \lambda_{\max} 10^{-\delta t/(T-1)}$ with a small $\delta$, say $\delta = 10^{-2}$ or $10^{-3}$, see for instance (Bühlmann and van de Geer, 2011)[2.12.1] or the *glmnet* package (Friedman et al., 2010). The parameter $\delta$ has an important influence on the computational burden: computing time tends to increase for small $\lambda$, the primal iterates being less and less sparse, and the problem to solve more and more ill-posed. It is customary to start from the largest regularizer $\lambda_0 = \lambda_{\max}$ and then to perform iteratively the computation of $\hat{\beta}^{(\lambda_t)}$ after the one of $\hat{\beta}^{(\lambda_{t-1})}$. This leads to computing the models in the order of increasing complexity: this allows important speed-up by benefiting of warm start strategies.

Here we propose a simple pathwise algorithm divided in two step:

- **Active warm start**: improve solver initialization by solving the problem restricted to an initial estimation of the support based on sequential informations along the regularization path (see Section 4.4 for details on the various strategies investigated).

- **Dynamic Gap Safe Screening**: use the informations gained during the iterations of the algorithm to obtain a smaller safe region therefore a greater elimination of inactive variables (see Section 4.3).

We summarize our strategy for solving the problem given by Equation (3) in Algorithm 1 and 2. The notation Solver (...) refers to any numerical solver that produces an approximation of the solution of (3) and SolverUpdate (...) is the updating scheme of the current vector along the iterations[5]. We consider solvers that can use a (primal) warm start point.

## 4. Screening strategies and theoretical analysis

We now describe the simplest safe rule strategy, which we refer to as the static strategy.

### 4.1 Static Safe Rules

The first static safe rule, introduced by El Ghaoui et al. (2012), discards variables before any computation thanks to Proposition 4. Here, the (safe) sphere is fixed once and for all,

---

5. For our experiments we have focused on (block) coordinate descent solvers

---

**Algorithm 1** Pathwise algorithm with active warm start

---

**Input** : $X$, $y$, $\epsilon$, $K$, $f^{ce}$, $(\lambda_t)_{t\in[T-1]}$

**for** $t \in [T-1]$ **do**

    $\beta = \breve{\beta}^{(\lambda_{t-1})}$ and                                                `// Get previous ε-solution`

    Get an initial (safe or not) support estimator $\mathcal{S} = \mathcal{S}(\breve{\beta}^{(\lambda_{t-1})})$

    $\beta_{\mathcal{S}} = \text{Solver}(X_{\mathcal{S}}, y, \beta_{\mathcal{S}}, \epsilon, K, f^{ce}, \lambda_t)$         `// Active warm start`

    $\breve{\beta}^{(\lambda_t)} = \text{Solver}(X, y, \beta, \epsilon, K, f^{ce}, \lambda_t)$         `// Solve over all variables`

**Output**: $\left(\breve{\beta}^{(\lambda_t)}\right)_{t\in[T-1]}$

---

**Algorithm 2** Iterative solver with GAP safe rules: Solver $(X, y, \beta, \epsilon, K, f^{ce}, \lambda)$

---

**Input** : $X$, $y$, $\beta$, $\epsilon$, $K$, $f^{ce}$, $\lambda$          `// Warm start is authorized here through β`

**for** $k \in [K]$ **do**

    **if** $k \bmod f^{ce} = 1$ **then**

        Compute a dual variable $\theta = -G(X\beta)/\max(\lambda, \Omega^D(X^\top G(X\beta)))$

        Stop if $\text{Gap}_\lambda(\beta, \theta) \leqslant \epsilon$

        $r = \sqrt{\frac{2\text{Gap}_\lambda(\beta,\theta)}{\gamma\lambda^2}}$         `// Get Gap Safe radius as in Equation (11)`

        $\mathcal{A} = \{g \in \mathcal{G} : \Omega_g^D(X_g^\top \theta) + r\Omega_g^D(X_g) \geqslant 1\}$   `// Get Safe active set as in Equation (12)`

    $\beta_{\mathcal{A}} = \text{SolverUpdate}(X_{\mathcal{A}}, y, \beta_{\mathcal{A}}, \lambda)$         `// Solve on current Safe active set`

**Output**: $\breve{\beta}^{(\lambda)}$

---

hence the name static. The static rule reads:

> <u>Static sphere rule:</u>    If    $\Omega_g^D(X_g^\top \theta_{\max}) + r_{\max}\Omega_g^D(X_g) < 1$,    then    $\hat{\beta}_g^{(\lambda)} = 0$ ,
>
>       Center:      $\theta_{\max} := -G(0)/\lambda_{\max}$ ,
>
>       Radius:      $r_{\max} := \sqrt{2\text{Gap}_\lambda(0, \theta_{\max})/\gamma\lambda^2}$ .

There is a threshold $\lambda_{\text{critic}}$ such that for any $\lambda$ smaller than $\lambda_{\text{critic}}$ the test from the Static sphere rule can never be satisfied. This phenomenon appears clearly in the numerical experiments presented in Section 6. In simple cases a closed form for $\lambda_{\text{critic}}$ can even be provided. For instance, in the case of the Group Lasso, El Ghaoui et al. (2012) proposed to use $r_{\max} = \left|\frac{1}{\lambda} - \frac{1}{\lambda_{\max}}\right| \|y\|_2$, and simple calculation gives:

$$\lambda_{\text{critic}} := \lambda_{\max} \times \min_{g\in\mathcal{G}} \frac{\|y\|_2 \, \Omega_g^D(X_g)}{\lambda_{\max} + \|y\|_2 \, \Omega_g^D(X_g) - \Omega_g^D(X_g G(0))} .$$

## 4.2 Sequential Safe Rules

Provided that the $\lambda$'s are close enough along the regularization parameters, knowing an estimate of $\hat{\beta}^{(\lambda_{t-1})}$ gives a clever initialization to compute $\hat{\beta}^{(\lambda_t)}$. To initialize the solver for a new $\lambda_t$, a natural choice is to set the primal variable equal to $\breve{\beta}^{(\lambda_{t-1})}$, an approximation of $\hat{\beta}^{(\lambda_{t-1})}$ output by the solver (at a prescribed precision). This popular strategy is referred to as "warm start" in the literature (Friedman et al., 2007). On top of this standard strategy, one can reuse prior dual information to improve the screening as well (El Ghaoui et al., 2012). This leads to the sequential strategy to screen for a new $\lambda_t$:

$$\underline{\text{Sequential sphere rule:}} \quad \text{If} \quad \Omega_g^D(X_g^\top \tilde{\theta}^{(\lambda_{t-1})}) + r_t \Omega_g^D(X_g) < 1, \quad \text{then} \quad \hat{\beta}_g^{(\lambda_t)} = 0 \;,$$

$$\text{Center:} \quad \tilde{\theta}^{(\lambda_{t-1})} := \Theta(-G(X\check{\beta}^{(\lambda_{t-1})})/\lambda_{t-1}) \;,$$

$$\text{Radius:} \quad r_t := \sqrt{2\mathrm{Gap}_{\lambda_t}(\check{\beta}^{(\lambda_{t-1})}, \tilde{\theta}^{(\lambda_{t-1})})/\gamma\lambda_t^2} \;.$$

**Remark 10** *Previous works in the literature (Wang et al., 2012; Wang and Ye, 2014; Wang et al., 2014; Lee and Xing, 2014) proposed sequential safe rules, though they were generally used in an unsafe way in practice. Indeed, such rules relied on the exact knowledge of $\hat{\theta}^{(\lambda_{t-1})}$ to screen out coordinates of $\hat{\beta}^{(\lambda_t)}$. Unfortunately, it is impossible to obtain such a point[6] since it is the solution of an optimization problem typically solved by an iterative solver, hence such it is only known up to a limited precision. By ignoring such inaccuracy in the knowledge of $\hat{\theta}^{(\lambda_{t-1})}$ one can wrongly eliminate variables that do belong to the support of $\hat{\beta}^{(\lambda_t)}$. Without a posteriori checking the screened out features, this could prevent the algorithm from converging, as shown in (Ndiaye et al., 2016b, Appendix B).*

### 4.3 Dynamic Safe Rules

Another road to speed up solvers using screening rules was proposed by Bonnefoy et al. (2014, 2015) under the name "dynamic safe rules". For a fixed $\lambda$, it consists in performing screening along with the iterations of the optimization algorithm used to solve Problem (3). Denoting by $k$ the iteration number, they introduced a rule for the Lasso that consists of a safe sphere with center $y/\lambda$ and radius $\|y/\lambda - \theta_k\|$, where $\theta_k$ is a current dual feasible point.

Let us consider a sequence $(\beta_k)$ that converges to a primal solution $\hat{\beta}^{(\lambda)}$. For creating a dual feasible point, we apply the rescaling introduced in Equation (9) to $z = -G(X\beta_k)/\lambda$ and the dynamic strategy can be summarized by

$$\underline{\text{Dynamic sphere rule:}} \quad \text{If} \quad \Omega_g^D(X_g^\top \theta_k) + r_k \Omega_g^D(X_g) < 1, \quad \text{then} \quad \hat{\beta}_g^{(\lambda)} = 0 \;,$$

$$\text{Center:} \quad \theta_k := \Theta(-G(X\beta_k)/\lambda) \text{ defined thanks to (9)} \;,$$

$$\text{Radius:} \quad r_k := \sqrt{2\mathrm{Gap}_\lambda(\beta_k, \theta_k)/\gamma\lambda^2} \;.$$

In practice the computation of the duality gap can be expensive due to the matrix vector operations needed to compute $X^\top G(X\beta_k)$. For instance in the Lasso case, a dual gap computation requires almost as much computation as a full pass of coordinate descent over the data. Hence, it is recommended to evaluate the dynamic (safe) rule only every few passes over the data set. In all our experiments, we have set this screening frequency parameter to $f^{ce} = 10$.

Note that contrary to the original dynamic screening rules proposed by Bonnefoy et al. (2014, 2015), the Gap Safe rules we introduced are converging in the sense that our safe regions converge to the singleton $\{\hat{\theta}^{(\lambda)}\}$ (see Section 4 for more details). Indeed, their proposed safe sphere was centered on $y/\lambda$, and their radius can only be greater than $\|y/\lambda - \hat{\theta}^{(\lambda)}\|$ in the Lasso case they consider. We provide a visual comparison in Figure 1.

---

6. Except for $\lambda_0 = \lambda_{\max}$, where $\theta_{\max} := -G(0)/\lambda_{\max}$ can be computed exactly

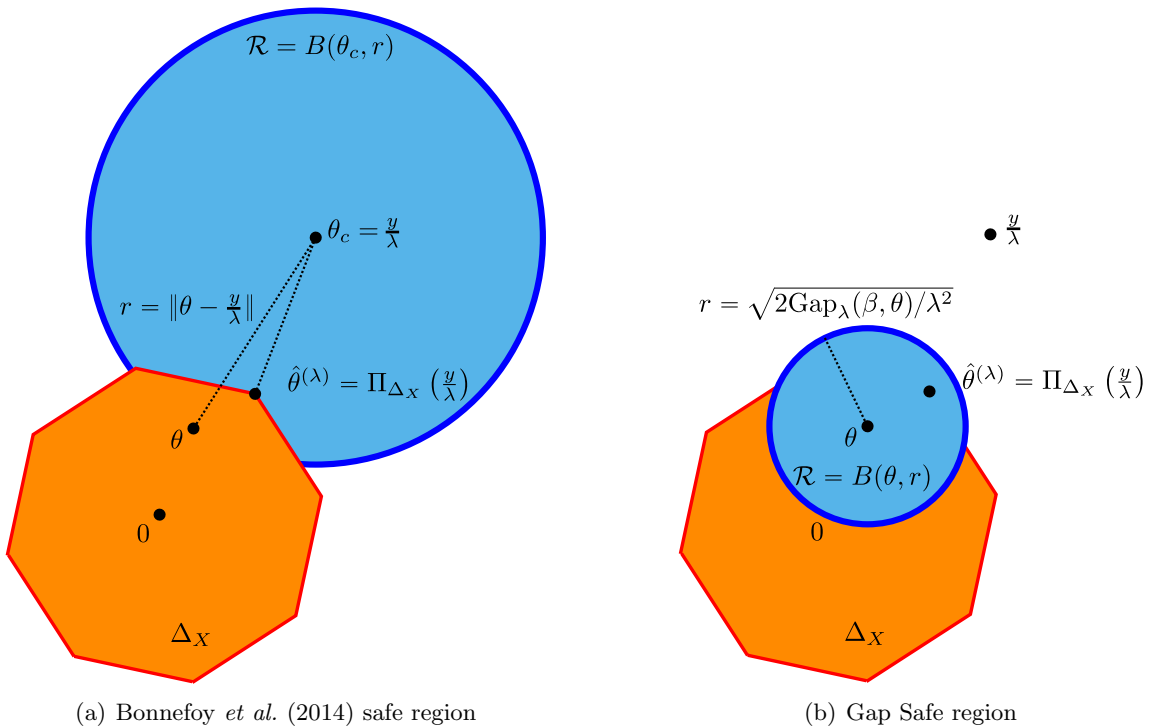

(a) Bonnefoy *et al.* (2014) safe region

(b) Gap Safe region

Figure 1: Illustration of safe region differences between Bonnefoy *et al.* (2014) and Gap Safe strategies for the Lasso case; note that $\gamma = 1$ in this case. Here $\beta$ is a primal point, $\theta$ is a dual feasible point (the feasible region $\Delta_X$ is in orange, while the respective safe balls $\mathcal{R}$ are in blue).

## 4.4 Active Warm Start

An another variant to further reduce running time in the *active warm start*, recently introduced by Ndiaye et al. (2016a) for speeding-up concomitant Lasso computations. Instead of simply leveraging the previous primal solution, the *active warm start* strategy also makes use of the previous safe active set $\mathcal{A}(\theta_{t-1}, r_{t-1})$, with $\theta_{t-1} = \check{\theta}^{(\lambda_{t-1})}$ and $r_{t-1} = r_{\lambda_{t-1}}(\check{\beta}^{(\lambda_{t-1})}, \check{\theta}^{(\lambda_{t-1})})$. The idea is to take as a new primal warm start point, the (approximate) minimizer of $P_{\lambda_t}$ under the additional constraint that its support is included in the safe active set $\mathcal{A}(\theta_{t-1}, r_{t-1})$ *i.e.*,

$$\widetilde{\beta}^{(\lambda_{t-1}, \lambda_t)} \in \underset{\beta \in \mathbb{R}^p}{\arg \min} \, F(\beta) + \lambda_t \Omega(\beta) \text{ s.t. } \text{supp}(\beta) \subseteq \mathcal{A}(\theta_{t-1}, r_{t-1}) \ . \tag{13}$$

In (13), we still choose $\check{\beta}^{(\lambda_{t-1})}$ as a standard warm start initialization with the same number of inner loops and/or accuracy as in (3) (to avoid the multiplication of parameters to be set by the user). Note that un-safe estimators of the active set can be used as for active warm start. In practice, we can use the (un-safe) strong active set provided by the Strong rules introduced by Tibshirani et al. (2012). This *Strong warm start* strategy is detailed in Section 4.6.

### 4.5 Theoretical Analysis

Dynamic safe screening rules have practical benefits since they increase the number of screened out variables as the algorithm proceeds. In this section, it is shown that Gap Safe rules allow to have sharper and sharper dual regions along the iterations, accelerating support identification. Before this, the following proposition states that if one relies on a primal converging algorithm, then the dual sequence we propose is also converging. Note that the convergence is maintained to the same primal solution when the primal solution is non-unique.

**Proposition 11 (Convergence of the Dual Points)** *Let $\beta_k$ be a current estimate of a primal solution $\hat{\beta}^{(\lambda)}$ and $\theta_k = \Theta(-G(X\beta_k)/\lambda)$ be the current estimate of $\hat{\theta}^{(\lambda)}$. Then, $\lim_{k\to+\infty} \beta_k = \hat{\beta}^{(\lambda)}$ implies $\lim_{k\to+\infty} \theta_k = \hat{\theta}^{(\lambda)}$.*

**Proof** Let $\alpha_k = \max(\lambda, \Omega^D(X^\top G(X\beta_k)))$, we have:

$$\left\| \theta_k - \hat{\theta}^{(\lambda)} \right\|_2 = \left\| \frac{G(X\hat{\beta}^{(\lambda)})}{\lambda} - \frac{G(X\beta_k)}{\alpha_k} \right\|_2 \leqslant \left| \frac{1}{\lambda} - \frac{1}{\alpha_k} \right| \|G(X\beta_k)\|_2 + \frac{\left\| G(X\hat{\beta}^{(\lambda)}) - G(X\beta_k) \right\|_2}{\lambda}.$$

If $\beta_k \to \hat{\beta}^{(\lambda)}$, then $\alpha_k \to \max(\lambda, \Omega^D(X^\top G(X\hat{\beta}^{(\lambda)}))) = \max(\lambda, \lambda\Omega^D(X^\top\hat{\theta}^{(\lambda)})) = \lambda$ since $G(X\hat{\beta}^{(\lambda)}) = -\lambda\hat{\theta}^{(\lambda)}$ thanks to the link-equation (5) and since $\hat{\theta}^{(\lambda)}$ is feasible, $\Omega^D(X^\top\hat{\theta}^{(\lambda)}) \leqslant 1$. Hence, both terms in the previous inequality converge to zero. ∎

Let us now describe the notion of converging safe regions and converging safe rules introduced in (Fercoq et al., 2015, Definition 1).

**Definition 12** *Let $(\mathcal{R}_k)_{k\in\mathbb{N}}$ be a sequence of closed convex sets in $\mathbb{R}^n$ containing $\hat{\theta}^{(\lambda)}$. It is a converging sequence of safe regions if the diameters of the sets converge to zero. The associated safe screening rules are referred to as converging.*

When $\theta_k = \Theta(-G(X\beta_k)/\lambda)$, Proposition 11 guarantees that Gap Safe spheres are converging. Indeed, the sequence of radius $r_k = (2\mathrm{Gap}_\lambda(\beta_k, \theta_k)/(\gamma\lambda^2))^{1/2}$ converges to 0 with $k$ by strong duality, hence the sequence $\mathcal{B}(\theta_k, r_k)$ converges to $\{\hat{\theta}^{(\lambda)}\}$ which means that the proposed Gap Safe sphere is asymptotically optimal.

We now prove that one recovers a specific set, called the *equicorrelation set* in finite time with any converging strategy.

**Definition 13** *The equicorrelation set is defined as $\mathcal{E}_\lambda := \left\{ g \in \mathcal{G} : \Omega_g^D(X_g^\top \hat{\theta}^{(\lambda)}) = 1 \right\}$.*

Indeed, the following proposition asserts that after a finite number of steps, the equicorrelation set $\mathcal{E}_\lambda$ is exactly identified. Such a property is sometimes referred to as finite identification of the support (Liang et al., 2014) and is summarized in the following proposition. Yet, note that the (primal) optimal support can be strictly smaller than the equicorrelation set, see Tibshirani (2013). For clarity, links between optimal support, sure active sets, equicorrelation set are illustrated in Figure 2.

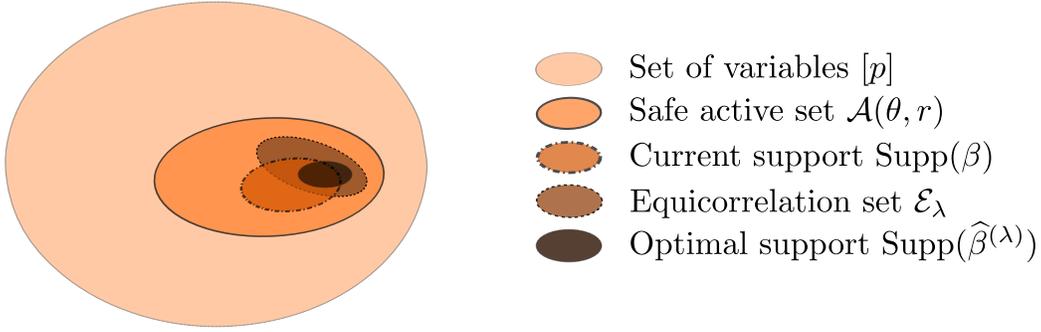

Set of variables $[p]$
Safe active set $\mathcal{A}(\theta, r)$
Current support $\text{Supp}(\beta)$
Equicorrelation set $\mathcal{E}_\lambda$
Optimal support $\text{Supp}(\widehat{\beta}^{(\lambda)})$

Figure 2: Illustration of the inclusions between several remarkable sets: $\text{supp}(\beta) \subseteq \mathcal{A}(\theta, r) \subseteq [p]$ and $\text{supp}(\widehat{\beta}^{(\lambda)}) \subseteq \mathcal{E}_\lambda \subseteq \mathcal{A}(\theta, r) \subseteq [p]$, where $\beta, \theta$ is a primal/dual pair.

**Proposition 14 (Identification of the Equicorrelation Set)**
*For any sequence of converging safe active set $(\mathcal{A}(\theta_k, r_k))_{k \in \mathbb{N}}$, we have $\lim_{k \to \infty} \mathcal{A}(\theta_k, r_k) = \mathcal{E}_\lambda$. More precisely, there exists an integer $k_0 \in \mathbb{N}$ such that $\mathcal{A}(\theta_k, r_k) = \mathcal{E}_\lambda$ for all $k \geqslant k_0$.*

**Proof** We proceed by double inclusion. First remark that $\mathcal{E}_\lambda = \mathcal{A}(\hat{\theta}^{(\lambda)}, \hat{r}_\lambda)$ where $\hat{r}_\lambda := 2(P_\lambda(\hat{\beta}^{(\lambda)}) - D_\lambda(\hat{\theta}^{(\lambda)}))/\gamma\lambda^2 = 0$ (thanks to strong duality), so for all $k \in \mathbb{N}$, we have $\mathcal{E}_\lambda \subseteq \mathcal{A}(\theta_k, r_k)$.

Reciprocally, suppose that there exists a non active group $g \in \mathcal{G}$ *i.e.*, $\Omega_g^D(X_g^\top \hat{\theta}^{(\lambda)}) < 1$ that remains in the active set $\mathcal{A}(\theta_k, r_k)$ for all iterations *i.e.*, $\forall k \in \mathbb{N}$, $\Omega_g^D(X_g^\top \theta_k) + r_k \Omega_g^D(X_g) \geqslant 1$. Since $\lim_{k \to \infty} \theta_k = \hat{\theta}^{(\lambda)}$ and $\lim_{k \to \infty} r_k = 0$, we obtain $\Omega_g^D(X_g^\top \hat{\theta}^{(\lambda)}) \geqslant 1$ by passing to the limit. Hence, by contradiction, there exits an integer $k_0 \in \mathbb{N}$ such that $\mathcal{E}_\lambda^c \subseteq \mathcal{A}(\theta_{k_0}, r_{k_0})^c$. ∎

### 4.6 Alternative Strategies: a Brief Survey

#### 4.6.1 The Seminal Safe Regions

The first Safe Screening rules introduced by El Ghaoui et al. (2012) can be generalized to Problem (3) as follows. Take $\hat{\theta}^{(\lambda_0)}$ the optimal solution of the dual problem (4) with a regularization parameter $\lambda_0$. Since $\hat{\theta}^{(\lambda)}$ is optimal for problem (4) one obtains $\hat{\theta}^{(\lambda)} \in \{\theta : D_\lambda(\theta) \geqslant D_\lambda(\hat{\theta}^{(\lambda_0)})\}$. This set was proposed as a safe region by El Ghaoui et al. (2012).

In the regression case (where $f_i(z) = (y_i - z)^2/2$), it is straightforward to see that it corresponds to the safe sphere $\mathcal{B}(y/\lambda, \|y/\lambda - \hat{\theta}^{(\lambda_0)}\|_2)$.

#### 4.6.2 ST3 and Dynamic ST3

Following (7), the safe sphere test given by Equation (8) is more efficient when $\theta$ is near $\hat{\theta}^{(\lambda)}$ and $r$ close to zero. This motivated the following improvements.

A refined sphere rule can be obtained in the regression case by exploiting geometric informations in the dual space. This method was originally proposed in Xiang et al. (2011) and extended in Bonnefoy et al. (2014) with a dynamic refinement of the safe region.

Let $g_\star \in \arg\max_{g \in \mathcal{G}} \Omega_g^D(X^\top y)$ (note that $\Omega_{g_\star}^D(X^\top y) = \lambda_{\max}$), and let us define

$$\mathcal{V}_\star := \{\theta \in \mathbb{R}^n : \Omega_{g_\star}^D(X_{g_\star}^\top \theta) \leqslant 1\} \text{ and } \mathcal{H}_\star := \{\theta \in \mathbb{R}^n : \Omega_{g_\star}^D(X_{g_\star}^\top \theta) = 1\}.$$

We assume that the dual norm is differentiable at $X_{g_\star}^\top y / \lambda_{\max}$ (which is true in all the cases presented in Section 5). Let $\eta := X_{g_\star} \nabla \Omega_{g_\star}^D(X_{g_\star}^\top y / \lambda_{\max})$ be the vector normal to $\mathcal{V}_\star$ at $y / \lambda_{\max}$ and define

$$\theta := \Pi_{\mathcal{H}_\star}\left(\frac{y}{\lambda}\right) = \frac{y}{\lambda} - \frac{\langle \frac{y}{\lambda}, \eta \rangle - 1}{\|\eta\|_2^2}\eta \text{ and } r_\theta := \sqrt{\left\|\frac{y}{\lambda} - \theta\right\|_2^2 - \left\|\frac{y}{\lambda} - \theta\right\|_2^2}\ ,$$

where $\theta \in \Delta_X$ is any dual feasible vector. Following the proof in (Ndiaye et al., 2016b, Appendix D), one can show that $\hat{\theta}^{(\lambda)} \in \mathcal{B}(\theta, r_\theta)$. The special case where $\theta = y / \lambda_{\max}$ corresponds to the original ST3 introduced in Xiang et al. (2011) for the Lasso. A further improvement can be obtained by choosing dynamically $\theta = \theta_k$ along the iterations of an algorithm, this strategy corresponding to DST3 introduced in Bonnefoy et al. (2014, 2015) for the Lasso and Group Lasso, and in Ndiaye et al. (2016b) for the Sparse-Group Lasso.

### 4.6.3 DUAL POLYTOPE PROJECTION

In the regression case, Wang et al. (2012) explore other geometric properties of the dual solution. Their method is based on the non-expansiveness of projection operators[7]. Indeed, for $\hat{\theta}^{(\lambda)}$ (resp. $\hat{\theta}^{(\lambda_0)})$) being optimal dual solution of (4) with parameter $\lambda$ (resp. $\lambda_0$), one has: $\|\hat{\theta}^{(\lambda)} - \hat{\theta}^{(\lambda_0)}\|_2 = \|\Pi_{\Delta_X}(y/\lambda) - \Pi_{\Delta_X}(y/\lambda_0)\|_2 \leqslant \|y/\lambda - y/\lambda_0\|_2$ and hence $\hat{\theta}^{(\lambda)} \in \mathcal{B}(\hat{\theta}^{(\lambda_0)}, \|y/\lambda - y/\lambda_0\|_2)$. Unfortunately, those regions are intractable since they required the *exact* knowledge of the optimal solution $\hat{\theta}^{(\lambda_0)}$ which is not available in practice (except for $\lambda_0 = \lambda_{\max}$). It may lead to un-safe screening rules as discussed in Remark 10.

**Remark 15** *The preceding spheres are mainly based on the fact that $\hat{\theta}^{(\lambda)} = \Pi_{\Delta_X}(y/\lambda)$ which is limited to the regression case. Thus, those methods are not appropriate for more general data fitting term which greatly reduces the scope of such rules.*

**Remark 16** *The radius of the regions above do not converge to zero even in the dynamic case (DST3), and the (fixed) center of the preceding sphere can be far from $\hat{\theta}^{(\lambda)}$ when $\lambda$ gets small. Thus, those regions are not converging and are irrelevant for dynamic screening.*

### 4.6.4 STRONG RULES

The Strong rules were introduced in Tibshirani et al. (2012) as a *heuristic* extension of the safe rules. It consists in relaxing the *safe* properties to discard features more aggressively, and can be formalized as follows. Assume that the gradient of the data fitting term $\nabla F$ is group-wise non-expansive w.r.t. the dual norm $\Omega_g^D(\cdot)$ along the regularization path *i.e.*, that for any $g \in \mathcal{G}$, any $\lambda > 0, \lambda' > 0$, $\Omega_g^D\left(\nabla_g F(\hat{\beta}^{(\lambda)}) - \nabla_g F(\hat{\beta}^{(\lambda')})\right) \leqslant |\lambda - \lambda'|$. When choosing two regularization parameters such that $\lambda < \lambda'$ one has:

$$\lambda \Omega_g^D\left(X_g^\top \hat{\theta}^{(\lambda)}\right) = \Omega_g^D\left(\nabla_g F(\hat{\beta}^{(\lambda)})\right) \leqslant \Omega_g^D\left(\nabla_g F(\hat{\beta}^{(\lambda')})\right) + \Omega_g^D\left(\nabla_g F(\hat{\beta}^{(\lambda)}) - \nabla_g F(\hat{\beta}^{(\lambda')})\right)$$

$$\leqslant \Omega_g^D\left(\nabla_g F(\hat{\beta}^{(\lambda')})\right) + |\lambda - \lambda'|$$

$$= \lambda' \Omega_g^D\left(X_g^\top \hat{\theta}^{(\lambda')}\right) + \lambda' - \lambda.$$

---

7. The authors also proved an enhanced version of this safe region by using the firm non-expansiveness of the projection operator.

Combining this with the screening rule (7), one obtains:

$$\Omega_g^D\left(X_g^\top \hat{\theta}^{(\lambda')}\right) < \frac{2\lambda - \lambda'}{\lambda'} \implies \Omega_g^D\left(X_g^\top \hat{\theta}^{(\lambda)}\right) < 1 \implies \hat{\beta}_g^{(\lambda)} = 0. \tag{14}$$

The set of variables not eliminated is called the *strong active set* and is defined as:

$$\mathcal{STG}(\hat{\theta}^{(\lambda')}, \lambda, \lambda') := \left\{ g \in \mathcal{G} : \Omega_g^D\left(X_g^\top \hat{\theta}^{(\lambda')}\right) \geqslant \frac{2\lambda - \lambda'}{\lambda'} \right\}. \tag{15}$$

Note that Strong rules are un-safe because the non-expansiveness condition on the (gradient of the) data fitting term is usually not satisfied without stronger assumptions on the design matrix $X$; see discussion in (Tibshirani et al., 2012, Section 3). It requires the exact knowledge of $\hat{\theta}^{(\lambda')}$ which is not available in practice. Using such rules, the authors advised to check the KKT condition[8] a posteriori, to avoid removing wrongly some features.

To overcome this limitation, we propose to use the strong active set $\mathcal{STG}(\hat{\theta}^{(\lambda_{t-1})}, \lambda_t, \lambda_{t-1})$ defined by Equation (15) for an active warm start strategy (cf. Section 4.4). We compare below this strategy with the one using $\mathcal{A}(\theta_{t-1}, r_{t-1})$ in Equation (13) as initial active set. A similar strategy is also used in the "big lasso" package by Zeng and Breheny (2017) as a hybrid screening strategy that *"alleviates the computational burden of KKT post-convergence checking for the strong rules by not checking features that can be safely eliminated"*. However, our warm start strategy (active or strong) does not require post-processing steps.

### 4.6.5 Correlation Based Rule

Previous works in statistics have proposed various model-based screening methods to select important variables. Those methods discard variables with small correlation between the features and response variables. For instance Sure Independence Screening (SIS) by Fan and Lv (2008) reads: for a chosen critical threshold $\gamma$ (such that the number of selected variables is smaller than a prescribed proportion of the features),

If $\Omega_g^D(X_g^\top y) < \gamma$ then remove $X_g$ from the problem.

---

8. The post-processing for the Lasso (with the notation from Section 5.1) adds back variables violating the approximated KKT conditions

$$KKT_\epsilon : \begin{cases} |X_j^\top \theta| \leqslant 1 + \epsilon, & \text{if } \beta_j = 0, \\ |X_j^\top \theta - \text{sign}(\beta_j)| \leqslant \epsilon, & \text{if } \beta_j \neq 0. \end{cases} \tag{16}$$

One can show that $\text{Gap}_\lambda(\beta, \theta) \leqslant (1 - \lambda/\alpha)^2 \|y - X\beta\|^2 / 2 + \lambda\epsilon \|\beta\|_1$ where $\alpha = \max(\lambda, \|X^\top(y - X\beta)\|_\infty)$. Hence choosing $\epsilon = \epsilon'/P_\lambda(\beta) - (1 - \lambda/\alpha)^2$ imply an $\epsilon'$-duality gap.

|  | Lasso | Multi-task regr. | Logistic regr. | Multinomial regr. |
|---|---|---|---|---|
| $f_i(z)$ | $\frac{(y_i-z)^2}{2}$ | $\frac{\|Y_i-z\|^2}{2}$ | $\log(1+\mathrm{e}^z)-y_i z$ | $\log\left(\sum_{k=1}^{q}\mathrm{e}^{z_k}\right)-\sum_{k=1}^{q}Y_{i,k}z_k$ |
| $f_i^*(u)$ | $\frac{(u+y_i)^2-y_i^2}{2}$ | $\frac{\|u+y_i\|^2-\|Y_i\|_2^2}{2}$ | $\mathrm{Nh}(u+y_i)$ | $\mathrm{NH}(u+Y_i)$ |
| $G(\theta)$ | $\theta-y$ | $\theta-Y$ | $\frac{-\mathrm{e}^\theta}{1+\mathrm{e}^\theta}-y$ | $\mathrm{RowNorm}(\mathrm{e}^\theta)-Y$ |
| $\gamma$ | 1 | 1 | 4 | 1 |

|  | $\ell_1$ | $\ell_1/\ell_2$ | $\ell_1+\ell_1/\ell_2$ | |
|---|---|---|---|---|
| $\Omega(\beta)$ | $\|\beta\|_1$ | $\displaystyle\sum_{g\in\mathcal{G}}\|\beta_g\|_2$ | $\displaystyle\tau\|\beta\|_1+(1-\tau)\sum_{g\in\mathcal{G}}w_g\|\beta_g\|_2$ | |
| $\Omega^D(\xi)$ | $\displaystyle\max_{j\in[p]}|\xi_j|$ | $\displaystyle\max_{g\in\mathcal{G}}\|\xi_g\|_2$ | $\displaystyle\max_{g\in\mathcal{G}}\frac{\|\xi_g\|_{\epsilon_g}}{\tau+(1-\tau)w_g}$, | $\displaystyle\epsilon_g:=\frac{(1-\tau)w_g}{\tau+(1-\tau)w_g}$ |

Table 1: Useful ingredients for computing Gap Safe rules. We have used lower case to indicate when the parameters are vectors or not (following the notation used in Section 5.5 and 5.6). The function RowNorm consists in normalizing a (non-negative) matrix row-wise, such that each row sums to one. The details for computing the $\epsilon$-norm $\|\cdot\|_{\epsilon_g}$ is given in Proposition 17.

It is a marginal oriented variable selection method and it is worth noting that SIS can be recast as a static sphere test in linear regression scenarios:

$$\text{If } \Omega_g^D(X_g^\top y) < \gamma\left(1 - r\Omega_g^D(X_g)\right) \text{ then } \hat\beta_g^{(\lambda)} = 0 \text{ (remove } X_g).$$

Other refinements can also be found in the literature such as iterative screening (ISIS) (Fan and Lv, 2008), that bears some similarities with dynamic sphere safe tests.

## 5. Gap Safe Rule for Popular Estimators

We now detail how our results apply to relevant supervised learning problems. A summary synthesizing the different learning task we are addressing is given in Table 1.

### 5.1 Lasso

For the Lasso estimator (Tibshirani, 1996), the data-fitting term is the standard least square, *i.e.*, $F(\beta) = \|y - X\beta\|_2^2/2 = \sum_{i=1}^n (y_i - x_i^\top\beta)^2/2$ (meaning that $f_i(z) = (y_i - z)^2/2$). The regularization term enforces sparsity at the feature level and is defined by

$$\Omega(\beta) = \|\beta\|_1 \quad\text{and}\quad \Omega^D(\xi) = \|\xi\|_\infty = \max_{j\in[p]}|\xi_j|.$$

### 5.2 Group Lasso

For the Group Lasso estimator (Yuan and Lin, 2006), the data-fitting term is the same $F(\beta) = \|y - X\beta\|_2^2/2$ but the penalty considered enforces group sparsity. Hence, we consider the norm $\Omega(\beta) = \Omega_w(\beta)$, often referred to as an $\ell_1/\ell_2$ norm, defined by:

$$\Omega_w(\beta) := \sum_{g \in \mathcal{G}} w_g \left\| \beta_g \right\|_2 \quad \text{and} \quad \Omega_w^D(\xi) := \max_{g \in \mathcal{G}} \frac{\left\| \xi_g \right\|_2}{w_g} \ ,$$

where $w = (w_g)_{g \in \mathcal{G}}$ are some weights satisfying $w_g > 0$ for all $g \in \mathcal{G}$.

### 5.3 Sparse-Group Lasso

In the Sparse-Group Lasso case, we also have $\beta \in \mathbb{R}^p$ and $F(\beta) = \|y - X\beta\|_2^2/2$ but the regularization $\Omega(\beta) = \Omega_{\tau,w}(\beta)$ is defined by

$$\Omega_{\tau,w}(\beta) := \tau \|\beta\|_1 + (1 - \tau) \sum_{g \in \mathcal{G}} w_g \left\| \beta_g \right\|_2 ,$$

for $\tau \in [0, 1], w = (w_g)_{g \in \mathcal{G}}$ with $w_g \geq 0$ for all $g \in \mathcal{G}$. Note that we recover the Lasso if $\tau = 1$, and the Group Lasso if $\tau = 0$; the case where $w_g = 0$ for some $g \in \mathcal{G}$ together with $\tau = 0$ is excluded ($\Omega_{\tau,w}$ is not a norm in such a case). This estimator was introduced by Simon et al. (2013) to enforce sparsity both at the feature and at the group level, and was used in different applications such as brain imaging in Gramfort et al. (2013) or in genomics in Peng et al. (2010). Other hierarchical norms have also been proposed in Sprechmann et al. (2010) or Jenatton et al. (2011) and could be handled in our framework modulo additional technical details.

For the Sparse-Group Lasso, the geometry of the dual feasible set $\Delta_X$ is more complex (*cf.* Figure 3 for a comparison w.r.t. Lasso and Group Lasso). As a consequence, additional geometrical insights are needed to derive efficient safe rules, especially to compute the dual norm required by Equation (9) and the computation of the safe screening rules given in (7).

We now introduce the $\epsilon$-norm (denoted $\|\cdot\|_\epsilon$) as it has a connection with the Sparse-Group Lasso norm $\Omega_{\tau,w}$. The $\epsilon$-norm was first proposed in Burdakov (1988) for other purposes (see also Burdakov and Merkulov (2001)). For any $\epsilon \in [0, 1]$ and any $x \in \mathbb{R}^d$, $\|x\|_\epsilon$ is defined as the unique nonnegative solution $\nu$ of the following equation (for $\epsilon = 0$, we define $\|x\|_{\epsilon=0} := \|x\|_\infty$):

$$\sum_{i=1}^{d} (|x_i| - (1 - \epsilon)\nu)_+^2 = (\epsilon \nu)^2. \tag{17}$$

Using soft-thresholding, this is equivalent to solve in $\nu$ the equation $\|\mathrm{ST}_{(1-\epsilon)\nu}(x)\|_2 = \epsilon \nu$. Moreover, its dual norm is given by; see (Burdakov and Merkulov, 2001, Equation (42)):

$$\|\xi\|_\epsilon^D = \epsilon \|\xi\|_2^D + (1 - \epsilon)\|\xi\|_\infty^D = \epsilon \|\xi\|_2 + (1 - \epsilon)\|\xi\|_1 \ . \tag{18}$$

This allows to express the Sparse-Group Lasso norm $\Omega_{\tau,w}$ using the dual $\epsilon$-norm. We now derive an explicit formulation for the dual norm of the Sparse-Group Lasso, originally proposed in (Ndiaye et al., 2016b, Prop. 4):

**Proposition 17** *For all groups $g$ in $\mathcal{G}$, let us introduce $\epsilon_g := \dfrac{(1-\tau)w_g}{\tau + (1-\tau)w_g}$. Then, the Sparse-Group Lasso norm satisfies the following properties: for any $\beta$ and $\xi$ in $\mathbb{R}^p$,*

$$\Omega_{\tau,w}(\beta) = \sum_{g \in \mathcal{G}} (\tau + (1-\tau)w_g)\|\beta_g\|_{\epsilon_g}^D \quad and \quad \Omega_{\tau,w}^D(\xi) = \max_{g \in \mathcal{G}} \frac{\|\xi_g\|_{\epsilon_g}}{\tau + (1-\tau)w_g}.$$

$$\mathcal{B}_{\Omega_{t,w}^D} = \left\{\xi \in \mathbb{R}^p : \forall g \in \mathcal{G}, \|\mathrm{ST}_\tau(\xi_g)\|_2 \leqslant (1-\tau)w_g\right\}.$$

$$\partial \Omega_{\tau,w}(\beta) = \left\{z \in \mathbb{R}^p : \forall g \in \mathcal{G}, z_g \in \tau \partial \|\cdot\|_1(\beta_g) + (1-\tau)w_g \partial \|\cdot\|_2(\beta_g)\right\}.$$

*Hence the dual feasible set is given by*

$$\Delta_X = \left\{\theta \in \mathbb{R}^n : \forall g \in \mathcal{G}, \|\mathrm{ST}_\tau\left(X_g^\top \theta\right)\|_2 \leqslant (1-\tau)w_g\right\}$$
$$= \left\{\theta \in \mathbb{R}^n : \forall g \in \mathcal{G}, \|X_g^\top \theta\|_{\epsilon_g} \leqslant \tau + (1-\tau)w_g\right\}.$$

**Remark 18** *Computing the dual norm of the Sparse-Group Lasso involves solving for each group $g \in \mathcal{G}$, a problem similar to the one given in (17) which has a quadratic complexity. To overcome this difficulty, an efficient algorithm relying on sorting techniques was proposed in (Ndiaye et al., 2016b, Prop. 5) to perform exact dual norm evaluation.*

The Sparse-Group Lasso benefits from two levels of screening: the safe rules can detect both group-wise zeros and coordinate-wise zeros in the remaining groups: for any group $g$ in $\mathcal{G}$ and any safe sphere $\mathcal{B}(\theta, r)$, Equation (7) and the sub-differential of the Sparse-Group Lasso norm in Proposition 17 give (a detailed proof is given in (Ndiaye et al., 2016b, Appendix C))

> ***Group level safe screening rule:*** $\displaystyle \max_{\theta \in \mathcal{B}(\theta,r)} \frac{\|X_g^\top \theta\|_{\epsilon_g}}{\tau + (1-\tau)w_g} < 1 \Rightarrow \hat{\beta}_g^{(\lambda)} = 0.$
>
> ***Feature level safe screening rule:*** $\displaystyle \forall j \in g, \max_{\theta \in \mathcal{B}(\theta,r)} |X_j^\top \theta| < \tau \Rightarrow \hat{\beta}_j^{(\lambda)} = 0.$

Noting that $\|\mathrm{ST}_\tau(x)\|_2 = (1-\tau)w_g \Longleftrightarrow \|x\|_{\epsilon_g} = \tau + (1-\tau)w_g$, the above screening test on the group level can be rewritten as

$$\max_{\theta \in \mathcal{B}(\theta,r)} \left\|\mathrm{ST}_\tau(X_g^\top \theta)\right\|_2 < (1-\tau)w_g \Rightarrow \hat{\beta}_g^{(\lambda)} = 0.$$

The advantage of this formulation is that one can easily derive a "tight" upper-bound of the non-convex optimization problem in the left hand side of the preceding test. Indeed, we have $\mathrm{ST}_\tau(x) = x - \Pi_{\tau \mathcal{B}_\infty}(x)$ which brings us finally into a geometric problem easier to solve. We recall from (Ndiaye et al., 2016b, Prop. 1) that for any center $\theta \in \Delta_X$, any group $g \in \mathcal{G}$ and any $j \in g$, we have the following upper-bound

$$\max_{\theta \in \mathcal{B}(\theta,r)} |X_j^\top \theta| \leqslant |X_j^\top \theta| + r \|X_j\|_2,$$

$$\max_{\theta \in \mathcal{B}(\theta,r)} \left\|\mathrm{ST}_\tau(X_g^\top \theta)\right\|_2 \leqslant T_g := \begin{cases} \left\|\mathrm{ST}_\tau(X_g^\top \theta)\right\|_2 + r \|X_g\|_2, & \text{if } \|X_g^\top \theta\|_\infty > \tau, \\ \left(\|X_g^\top \theta\|_\infty + r \|X_g\|_2 - \tau\right)_+, & \text{otherwise.} \end{cases}$$

Hence we derive the two level of safe screening rule:

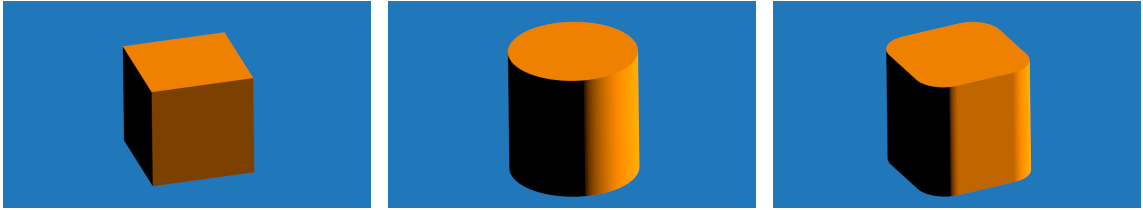

(a) Lasso dual ball $\mathcal{B}_{\Omega^D}$ for $\Omega^D(\theta) = \|\theta\|_\infty$.

(b) Group Lasso dual ball $\mathcal{B}_{\Omega^D}$ for $\Omega^D(\theta) = \max(\sqrt{\theta_1^2 + \theta_2^2}, |\theta_3|)$.

(c) Sparse-Group Lasso dual ball $\mathcal{B}_{\Omega^D} = \{\theta : \forall g \in \mathcal{G}, \|\mathrm{ST}_\tau(\theta_g)\|_2 \leqslant (1-\tau)w_g\}$.

Figure 3: Lasso, Group Lasso and Sparse-Group Lasso dual unit balls: $\mathcal{B}_{\Omega^D} = \{\theta : \Omega^D(\theta) \leqslant 1\}$. For the illustration, the group structure is chosen such that $\mathcal{G} = \{\{1,2\},\{3\}\}$, $i.e.,$ $g_1 = \{1,2\}, g_2 = \{3\}$, $n = p = 3$, $w_{g_1} = w_{g_2} = 1$ and $\tau = 1/2$.

**Proposition 19 (Safe screening rule for the Sparse-Group Lasso)**

> **Group level screening:** $\quad\quad \forall g \in \mathcal{G}, \quad$ if $T_g < (1-\tau)w_g, \quad\quad$ then $\hat{\beta}_g^{(\lambda)} = 0$.
>
> **Feature level screening:** $\quad \forall g \in \mathcal{G}, \forall j \in g$, if $|X_j^\top \theta| + r\,\|X_j\|_2 < \tau, \quad$ then $\hat{\beta}_j^{(\lambda)} = 0$.

In the same spirit than Proposition 14, for any safe region $\mathcal{R}$, $i.e.,$ a set containing $\hat{\theta}^{(\lambda)}$, we define two levels of active sets, one for the group level and one for the feature level:

$$\mathcal{A}_{\mathrm{gp}}(\mathcal{R}) := \{g \in \mathcal{G}, \max_{\theta \in \mathcal{R}} \left\|\mathrm{ST}_\tau(X_g^\top \theta)\right\|_2 \geqslant (1-\tau)w_g\},$$

$$\mathcal{A}_{\mathrm{ft}}(\mathcal{R}) := \bigcup_{g \in \mathcal{A}_{\mathrm{gp}}(\mathcal{R})} \{j \in g : \max_{\theta \in \mathcal{R}} |X_j^\top \theta| \geqslant \tau\}.$$

If one considers sequence of converging regions, then the next proposition (see (Ndiaye et al., 2016b, Prop. 3)) states that we can identify in finite time the optimal active sets defined as follows:

$$\mathcal{E}_{\mathrm{gp}} := \left\{g \in \mathcal{G} : \left\|\mathrm{ST}_\tau(X_g^\top \hat{\theta}^{(\lambda)})\right\|_2 = (1-\tau)w_g\right\}, \quad \mathcal{E}_{\mathrm{ft}} := \bigcup_{g \in \mathcal{E}_{\mathrm{gp}}} \left\{j \in g : |X_j^\top \hat{\theta}^{(\lambda)}| \geqslant \tau\right\}.$$

**Proposition 20** *Let* $(\mathcal{R}_k)_{k \in \mathbb{N}}$ *be a sequence of safe regions whose diameters converge to 0. Then,* $\lim_{k \to \infty} \mathcal{A}_{gp}(\mathcal{R}_k) = \mathcal{E}_{gp}$ *and* $\lim_{k \to \infty} \mathcal{A}_{ft}(\mathcal{R}_k) = \mathcal{E}_{ft}$.

## 5.4 $\ell_1$ Regularized Logistic Regression

Here, we consider the formulation given in (Bühlmann and van de Geer, 2011, Chapter 3) for the two-class logistic regression. In such a context, one observes for each $i \in [n]$ a class label $l_i \in \{1, 2\}$. This information can be recast as $y_i = \mathbb{1}_{\{l_i=1\}}$ (where $\mathbb{1}$ is the indicator function), and it is then customary to minimize (3) where

$$F(\beta) = \sum_{i=1}^n \left(-y_i x_i^\top \beta + \log\left(1 + \exp\left(x_i^\top \beta\right)\right)\right), \tag{19}$$

with $f_i(z) = -y_i z + \log(1 + \exp(z))$, and the penalty is simply the $\ell_1$ norm: $\Omega(\beta) = \|\beta\|_1$. Let us introduce Nh, the (binary) negative entropy function defined by:

$$\mathrm{Nh}(x) = \begin{cases} x \log(x) + (1-x) \log(1-x), & \text{if } x \in [0,1] \ , \\ +\infty, & \text{otherwise} \ . \end{cases} \tag{20}$$

We use the convention $0 \log(0) = 0$, and one can check that $f_i^*(z_i) = \mathrm{Nh}(z_i + y_i)$ and $\gamma = 4$.

**Remark 21** *We have privileged the formulation with the label $y \in \{0,1\}^n$ instead of $y \in \{+1, -1\}^n$ in order to be consistent with the multinomial cases below. One can simply switch from one formulation to the other thanks to the mapping $\tilde{y} = 2y - 1$.*

### 5.5 $\ell_1/\ell_2$ Multi-task Regression

The multi-task Lasso is a regression problem where the parameters form a matrix $\mathrm{B} \in \mathbb{R}^{p \times q}$. Denoting $n$ the number of observations for each task $k \in [q]$, it is defined as

$$\min_{\mathrm{B} \in \mathbb{R}^{p \times q}} \frac{1}{2} \|Y - X\mathrm{B}\|_F^2 + \lambda \sum_{j=1}^p \|\mathrm{B}_{j,:}\|_2 \,, \tag{21}$$

where $X \in \mathbb{R}^{n \times p}$ and $Y \in \mathbb{R}^{n \times q}$. Here we assume that the explanatory variables $X$ are shared among the tasks however the Gap Safe rules would readily apply to the non-shared design formulation as in Lee et al. (2010) or in Liu et al. (2009) since the loss is still smooth (*cf.* Remark 7).

Introducing the vec operator that vectorizes a matrix by stacking its columns to form a column vector, and the Kronecker product $\otimes$ of two matrices, the multi-task Lasso can be rewritten as a special case of Group Lasso. In fact, we have $n$ class of observations $c_i = (i + (k-1)n)_{k \in [q]}$ of size $q$ for each $i \in [n]$ (the overall number of observations is $n' = nq$) and $p$ groups $g_j = (j + (k-1)p)_{k \in [q]}$ such that $|g_j| = q$ for $j \in [p]$. The design matrix $\tilde{X} = I_q \otimes X \in \mathbb{R}^{n' \times p'} = \mathbb{R}^{nq \times pq}$ is a $q$-block diagonal matrix defined as $\tilde{X} = \mathrm{diag}(X, \ldots, X)$, $y = \mathrm{vec}(Y)$ and $\beta = \mathrm{vec}(\mathrm{B})$, we have:

$$\min_{\beta \in \mathbb{R}^{p'}} \frac{1}{2} \sum_{i=1}^n \|y_{c_i} - \tilde{x}_i^\top \beta\|_2^2 + \lambda \sum_{j=1}^p \|\beta_{g_j}\|_2 \,, \tag{22}$$

*i.e.*, $f_i(z) = \|y_{c_i} - z\|_2^2 / 2$. The advantage of this formulation is that it can be concisely written using the matrix forms of $y$ and $\beta$, without the need to actually construct the large matrix $X'$. This is particularly appealing for the implementation.

In signal processing, this model is also referred to as the Multiple Measurement Vector (MMV) problem. It allows to jointly select the same features for multiple regression tasks, see (Argyriou et al., 2006, 2008; Obozinski et al., 2010). This estimator has been used in various applications such as prediction of the location of a protein within a cell (Xu et al., 2011) or in neuroscience (Gramfort et al., 2012), for instance to diagnose Alzheimer's disease (Zhang et al., 2012).

### 5.6 $\ell_1/\ell_2$ Multinomial Logistic Regression

We adapt the formulation given in (Bühlmann and van de Geer, 2011, Chapter 3) for the multinomial regression. In such a context, one observes for each $i \in [n]$ a class label $l_i \in [q]$. This information can be recast into a matrix $Y \in \mathbb{R}^{n \times q}$ filled by 0's and 1's: $Y_{i,k} = \mathbb{1}_{\{l_i = k\}}$ (where $\mathbb{1}$ is the indicator function). In the same spirit as for the multi-task Lasso, a matrix $\mathrm{B} \in \mathbb{R}^{p \times q}$ is formed by $q$ vectors encoding the hyperplanes for the linear classification. Thus the multinomial $\ell_1/\ell_2$ regularized regression reads:

$$\min_{\mathrm{B} \in \mathbb{R}^{p \times q}} \sum_{i=1}^{n} \left( \sum_{k=1}^{q} -Y_{i,k} x_i^\top \mathrm{B}_{:,k} + \log \left( \sum_{k=1}^{q} \exp \left( x_i^\top \mathrm{B}_{:,k} \right) \right) \right) + \lambda \sum_{j=1}^{p} \|\mathrm{B}_{j,:}\|_2 . \qquad (23)$$

Using a similar reformulation as in Section 5.5 *i.e.*, defining $c_i = (i + (k-1)n)_{k \in [q]}$ for each $i \in [n]$ and $g_j = (j + (k-1)p)_{k \in [q]}$ for each $j \in [p]$, the $\ell_1/\ell_2$ multinomial logistic regression can be cast into our framework as:

$$\min_{\beta \in \mathbb{R}^{p'}} \sum_{i=1}^{n} f_i \left( \tilde{x}_i^\top \beta \right) + \lambda \sum_{j=1}^{p} \left\| \beta_{g_j} \right\|_2 , \qquad (24)$$

with $f_i : \mathbb{R}^q \to \mathbb{R}$ such that $f_i(z) = -y_{c_i}^\top z + \log \left( \sum_{k=1}^{q} \exp (z_k) \right)$. Note that generalizing (3) to functions $f_i : \mathbb{R}^q \to \mathbb{R}$ does not bear difficulties, see Ndiaye et al. (2015).

Let us introduce NH, the negative entropy function defined by

$$\mathrm{NH}(x) = \begin{cases} \sum_{i=1}^{q} x_i \log(x_i), & \text{if } x \in \Sigma_q = \{x \in \mathbb{R}_+^q : \sum_{i=1}^q x_i = 1\}, \\ +\infty, & \text{otherwise.} \end{cases} \qquad (25)$$

We use the convention $0 \log(0) = 0$, and one can check that $f_i^*(z) = \mathrm{NH}(z + Y_i)$ and $\gamma = 1$.

For multinomial logistic regression, $D_\lambda$ implicitly encodes the additional constraint $\theta \in \mathrm{dom}\, D_\lambda = \{\theta' \in \mathbb{R}^n : \forall i \in [n], -\lambda \theta'_{c_i} + y_{c_i} \in \Sigma_q\}$ where $\Sigma_q$ is the $q$ dimensional simplex, see Equation (25). By the dual scaling Equation (9), we have:

$$\theta = \Theta \left( \frac{-G(\tilde{X}\beta)}{\lambda} \right) = \frac{R}{\max(\lambda, \Omega^D(X^\top R))}, \text{ with } R = y - \mathrm{RowNorm}(\exp(\tilde{X}\beta))$$

where the function RowNorm consists in normalizing a (non-negative) matrix row-wise, such that each row sums to one. Thus for any $i \in [n]$ and $\alpha := \lambda/\max(\lambda, \Omega^D(X^\top R)) \in [0, 1]$,

$$-\lambda \theta_{c_i} + y_{c_i} = (1 - \alpha) y_{c_i} + \alpha \,\mathrm{RowNorm}(\exp(\tilde{x}_i^\top \beta))$$

which is a convex combination of elements in $\Sigma_q$. Hence the dual scaling (9) preserves this additional constraint.

**Remark 22** *The intercept has been neglected in our models for simplicity. The Gap Safe framework can also handle such a feature to the cost of more technical details (by adapting the results from Koh et al. (2007) for instance). However, in practice, the intercept can be handled in the present formulation by adding a constant column to the design matrix $X$. The intercept is then regularized. However, if the constant is set high enough, regularization is small and experiments show that it has little to no impact for high-dimensional problems. This is the strategy used in the* Liblinear *package by Fan et al. (2008). Another alternative could be to handle the constant term as is performed by El Ghaoui et al. (2012).*

## 6. Experiments

In this section we present results obtained with the Gap Safe rules on various data sets. Implementation[9] has been done in Python and Cython (Behnel et al., 2011) for low level critical parts. A coordinate descent algorithm is used with a scaled dual gap stopping criterion *i.e.*, we normalize the targeted accuracy $\epsilon$ (in the stopping criterion) in order to have a running time that is independent from the data scaling, *i.e.*, $\epsilon \leftarrow \epsilon \|y\|_2^2$ for the regression cases and $\epsilon \leftarrow \epsilon \min(n_1, n_2)/n$ where $n_i$ is the number of observations in the class $i$, for the logistic cases.

Note that in the Lasso case, to compare our method with the un-safe *strong rules* by Tibshirani et al. (2012) and with the sequential screening rule such as the *eddp+* by Wang et al. (2012), we have added an approximated KKT post-processing step. We do this following Footnote 8, since they require the previous (exact) dual optimal solution which is not available in practice (Ndiaye et al., 2016b, Appendix B). The same limit holds true for the *TLFre* approach of Wang and Ye (2014) addressing the Sparse-Group Lasso formulation, as well as for the method explored by Lee and Xing (2014) to handle overlapping groups and *slores* by Wang et al. (2014) for the binary logistic regression.

We have compared our method to various known safe screening rules (El Ghaoui et al., 2012; Xiang et al., 2011; Bonnefoy et al., 2014). For the Sparse-Group Lasso, such rules did not exist, so we have proposed natural extensions (Ndiaye et al., 2016b) thanks to exact computation of the dual norm in Proposition 17. For the Lasso estimator, we have also compared our implementation with the *Blitz* algorithm (Johnson and Guestrin, 2015) which combines Gap Safe screening rules, *Prox-Newton* coordinate descent and an active set strategy.

### 6.1 $\ell_1$ Lasso Regression

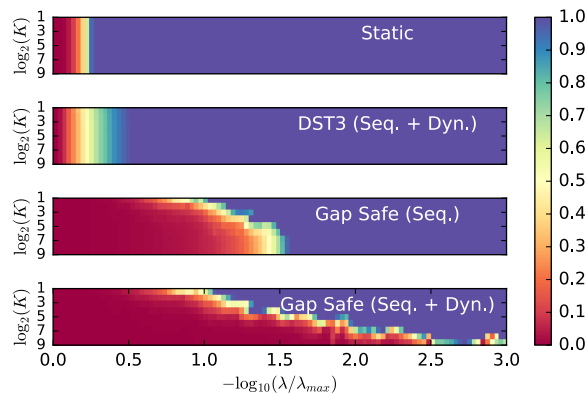

Figure 4: Lasso on the Leukemia (dense data with $n = 72$ observations and $p = 7129$ features). fraction of the variables that are active. Each line corresponds to a fixed number of iterations for which the algorithm is run.

---

9. The source code can be found in `https://github.com/EugeneNdiaye`.

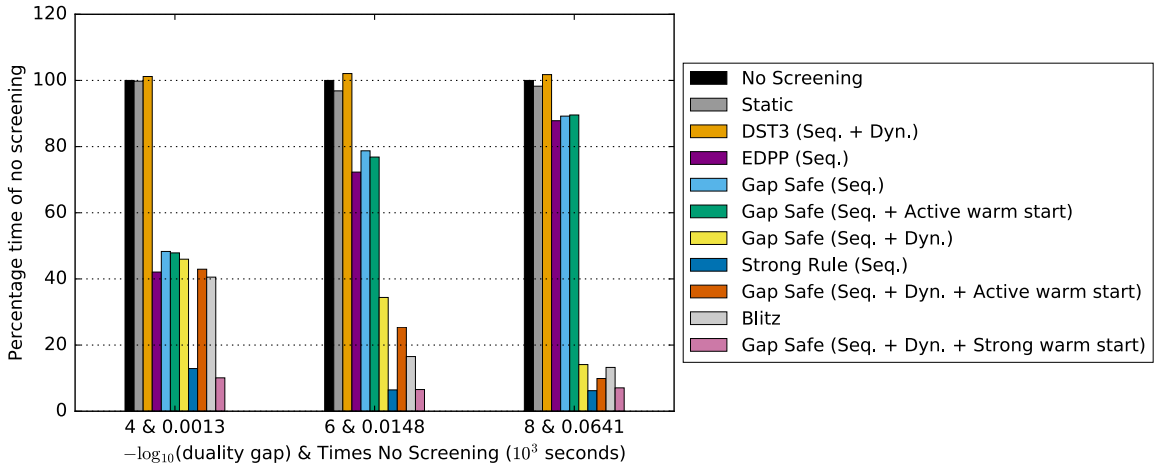

(a) Dense grid with 100 values of $\lambda$.

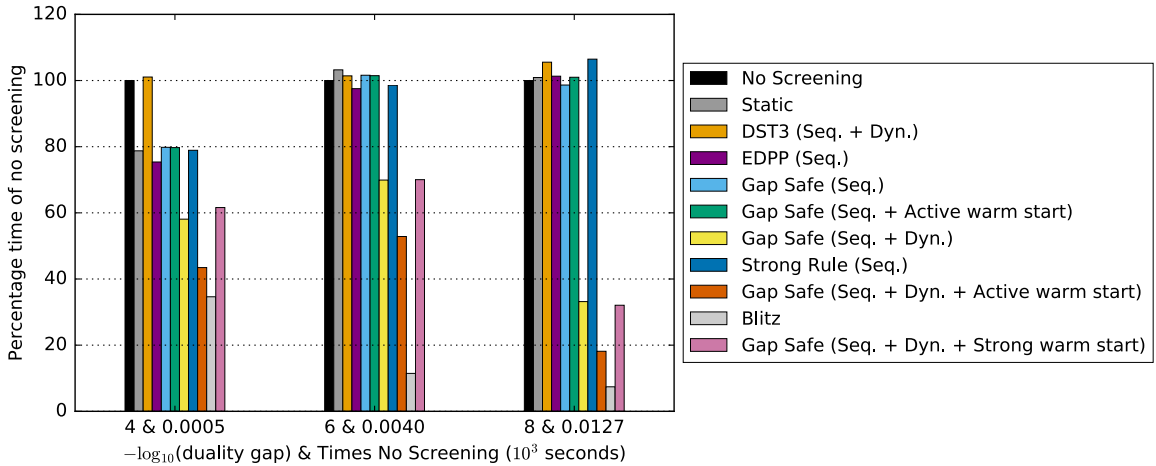

(b) Coarse grid with 10 values of $\lambda$.

Figure 5: Lasso on the Leukemia (dense data with $n = 72$ observations and $p = 7129$ features). Computation times needed to solve the Lasso regression path to desired accuracy for a grid of $\lambda$ from $\lambda_{\max}$ to $\lambda_{\max}/10^3$.

We have evaluated the computing time for the Gap Safe rules with and without active warm start, and compared with the static rule El Ghaoui et al. (2012) and the refined dynamic rule DST3 by Xiang et al. (2011), as well as Bonnefoy et al. (2015). We used the classic dense data set Leukemia, and the large sparse financial data set E2006-log1p available from LIBSVM[10]. We have normalized the column of $X$ and standardized $y$ to have zero mean and unit variance.

The experiments on Figure 4 focuses on the Leukemia data set. The screening performance for a fixed number of iterations, from 2 to $2^9$, is investigated for each $\lambda$. It demonstrates that increasing the number of iterations benefits to the dynamic screening rule. Also, the closer the estimate is from the global minimum, the better the screening. This is inline with the results in running time in the benchmark on Figure 5(a). Note that the dynamic Gap Safe rule is the only rule that significantly improves the running time of the Lasso.

---

10. `http://www.csie.ntu.edu.tw/~cjlin/libsvmtools/datasets/`

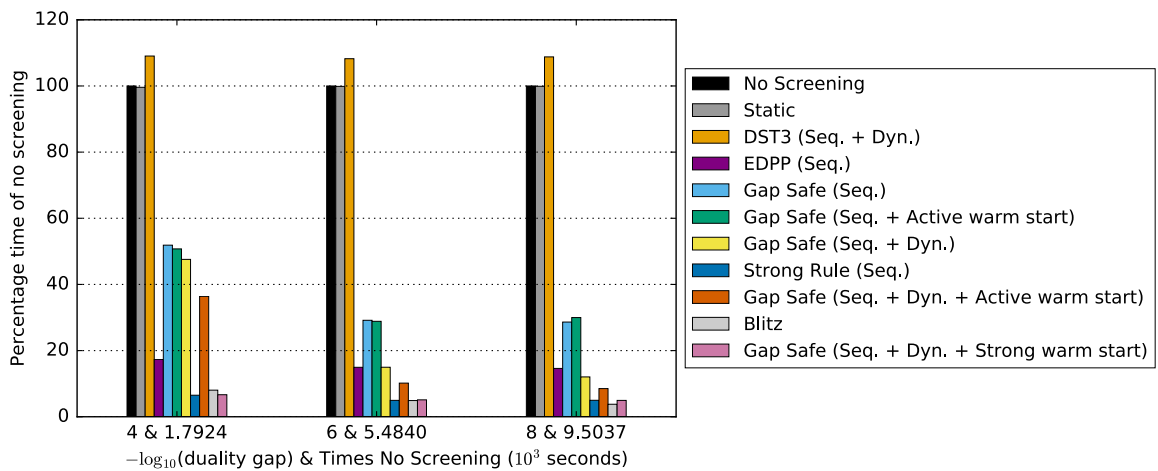

(a) Dense grid with 100 values of $\lambda$.

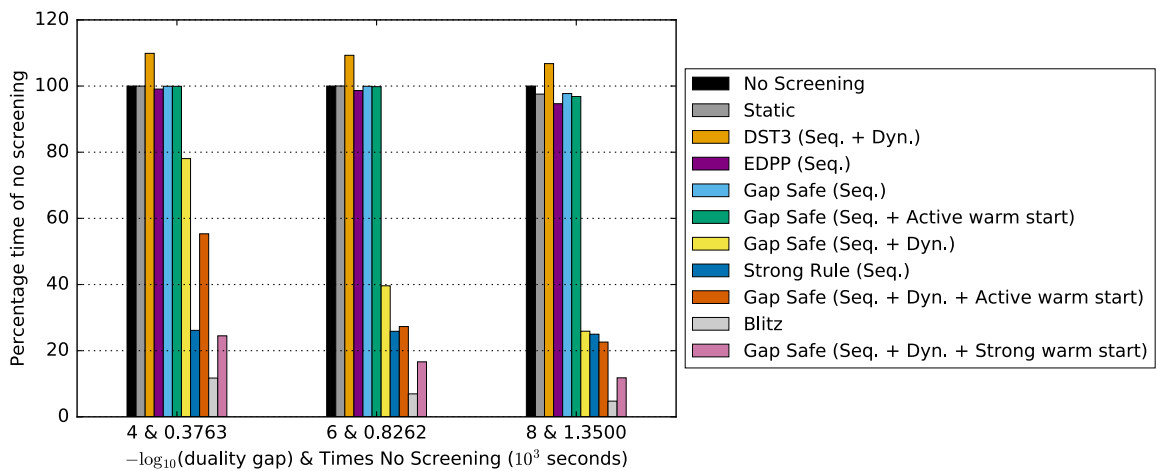

(b) Sparse grid with 10 values of $\lambda$.

Figure 6: Lasso on financial data E2006-log1p (sparse data with $n = 16087$ observations and $p = 1668737$ features). Computation times needed to solve the Lasso regression path to desired accuracy for a grid of $\lambda$ from $\lambda_{\max}$ to $\lambda_{\max}/20$.

Results presented in the financial data set in Figure 6 are inline with the results on Leukemia. We observe that the *Blitz* algorithm (Johnson and Guestrin, 2015), also achieves a significant speed-up with gains in the same order of magnitude than our dynamic Gap Safe implementation combined with active or strong warm start. One advantage of our approach though, is the simplicity to insert it in any iterative algorithm as shown in Algorithm 1 and 2.

To demonstrate the limitations of the strong rules, we report in Figure 5(b) results with a coarse grid with only 10 values of $\lambda$ from $\lambda_{\max}$ to $\lambda_{\max}/10^3$ such that $2\lambda_t < \lambda_{t-1}$. The strong rules become then useless since the screening test (14) selects all variables, *i.e.*, $\mathcal{STG}(\hat{\theta}^{(\lambda_{t-1})}, \lambda_t, \lambda_{t-1}) = \mathcal{G}$. Overall, the greater the gap between grid points, the lower the benefits of (active) warm start.

In the experiment in Figure 6(b), we have stopped the grid at $\lambda_{\max}/20$ leading to a sparse solution with 1562 active variables. We obtain an important speed-up for both coarse and dense grids demonstrating the consistent efficiency of the active warm start strategy specially in a sparse regime.

Finally, with an extremely coarse grid, we therefore recommend the active warm start with the previous safe active set (which performance is only affected through the initialization point) rather than the strong active set (*cf.* Figure 5(b)).

## 6.2 $\ell_1$ Binary Logistic Regression

Results on the Leukemia data set for standard logistic regression are reported in Figure 7. We compare the dynamic strategy of Gap Safe to the sequential strategy. Results demonstrate the clear benefit of the dynamic rule in terms of high number of screened out variables. This is reflected in the graph of running times, which shows that dynamic Gap Safe rule with strong warm start can yield up to a $30\times$ speed-up compared to sequential rule and even more compared to an absence of screening (up to $50\times$ speed-up).

## 6.3 $\ell_1/\ell_2$ Multi-task Regression

To demonstrate the benefit of the Gap Safe screening rules for a multi-task Lasso problem we have considered neuroimaging data. Electroencephalography (EEG) and magnetoencephalography (MEG) are brain imaging modalities that allow to identify active brain regions. The problem to solve is a multi-task regression problem with squared loss where every task corresponds to a time instant. Using a multi-task Lasso one can constrain the recovered sources to be identical during a short time interval (Gramfort et al., 2012). This corresponds to a temporal stationary assumption. In this experiment we used a joint MEG/EEG data with 301 MEG and 59 EEG sensors leading to $n = 360$. The number of possible sources is $p = 22,494$ and the number of time instants is $q = 20$. With a $1\,\mathrm{kHz}$ sampling rate it is equivalent to say that the sources stay the same for $20\,\mathrm{ms}$.

Results are presented in Figure 8. The Gap Safe rule is compared with the dynamic safe rule from Bonnefoy et al. (2015). Figure 8(a) shows the fraction of active variables. It demonstrates that the Gap Safe rule screens out much more variables than the compared methods. Thanks to the converging nature of our rule, the more iterations are performed the more variables are screened out. On Figure 8(b), the computation time confirms the effective speed-up. Our rule significantly improves the computation time for all duality gap tolerance from $10^{-2}$ to $10^{-8}$, especially when accurate estimates are required, *e.g.*, for feature selection.

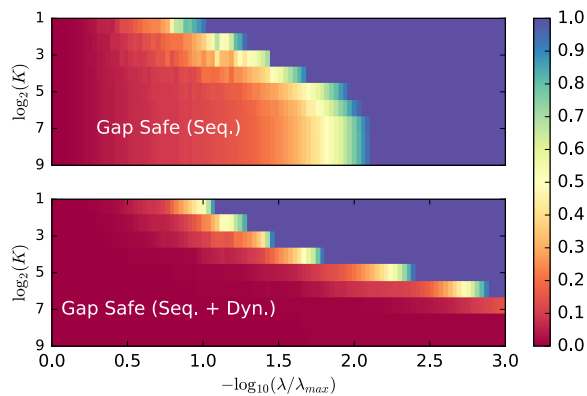

(a) Fraction of the variables that are active. Each line corresponds to a fixed number of iterations for which the algorithm is run.

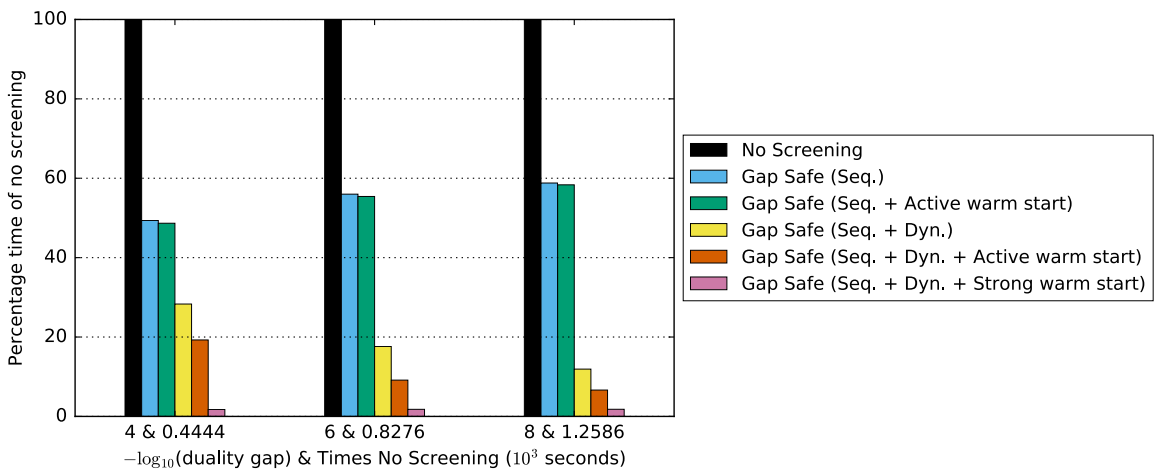

(b) Computation times needed to solve the logistic regression path to desired accuracy with 100 values of $\lambda$ from $\lambda_{max}$ to $\lambda_{max}/10^3$.

Figure 7: $\ell_1$ regularized binary logistic regression on the Leukemia (dense data with $n = 72$ observations and $p = 7129$ features). Sequential and full dynamic screening Gap Safe rules are compared.

## 6.4 Sparse-Group Lasso Regression

We consider the data set NCEP/NCAR Reanalysis 1 Kalnay et al. (1996) which contains monthly means of climate data measurements spread across the globe in a grid of $2.5° \times 2.5°$ resolutions (longitude and latitude $144 \times 73$) from 1948/1/1 to 2015/10/31. Each grid point constitutes a group of 7 predictive variables (*Air Temperature, Precipitable water, Relative humidity, Pressure, Sea Level Pressure, Horizontal Wind Speed* and *Vertical Wind Speed*) whose concatenation across time constitutes our design matrix $X \in \mathbb{R}^{814 \times 73577}$. Such data have therefore a natural group structure, with seven features per group. As target variable $y \in \mathbb{R}^{814}$, we use the values of *Air Temperature* in a neighborhood of Dakar.

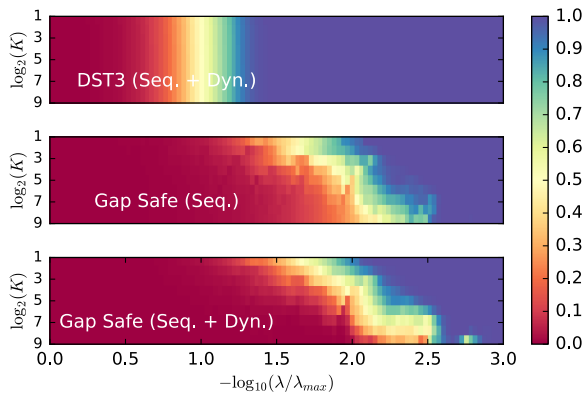

(a) Fraction of active variables as a function of $\lambda$ and the number of iterations $K$. The Gap Safe strategy has a much longer range of $\lambda$ with (red) small active sets.

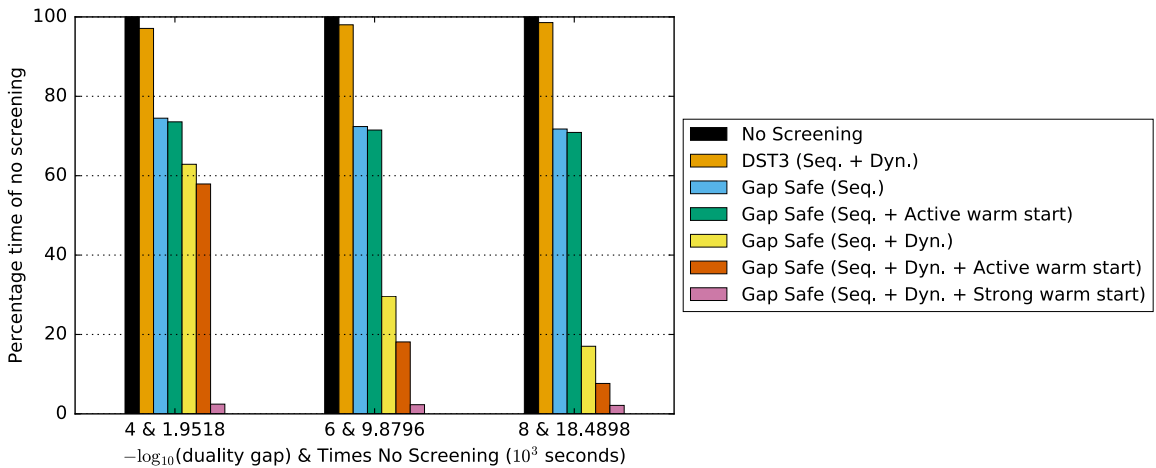

(b) Computation time to reach convergence using different screening strategies. We have run the algorithm with 100 values of $\lambda$ from $\lambda_{\max}$ to $\lambda_{\max}/10^3$.

Figure 8: Experiments on MEG/EEG brain imaging data set (dense data with $n = 360$ observations, $p = 22494$ features and $q = 20$ time instants).

For preprocessing, we remove the seasonality (we center the data month by month) and the trend (we remove the linear trend obtained by least squares) present in the data set. We then standardize the data so that each feature has a variance of one. This preprocessing is usually done in climate analysis to prevent some bias in the regression estimates. Similar data have been used in the past by Chatterjee et al. (2012), demonstrating that the Sparse-Group Lasso estimator is well suited for prediction in such climatology applications. Indeed, thanks to the sparsity structure, the estimates delineate via their support some predictive regions at the group level, as well as predictive feature via coordinate-wise screening.

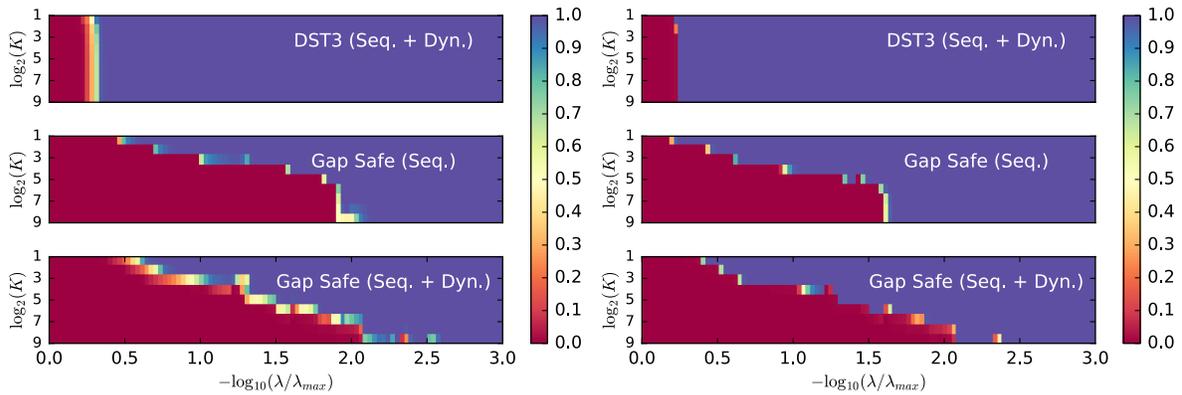

(a) Proportion of active coordinate-wise variables as a function of parameters ($\lambda_t$) and the number of iterations $K$.

(b) Proportion of active group variables as a function of parameters ($\lambda_t$) and the number of iterations $K$.

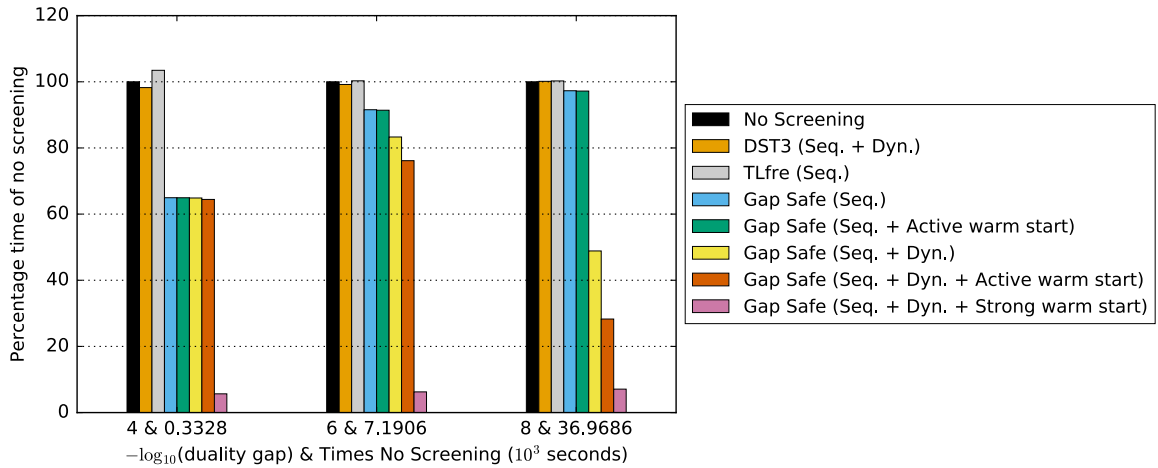

(c) Time to reach convergence as a function of increasing prescribed accuracy, using various screening strategies and a logarithmic grid from $\lambda_{\max}$ to $\lambda_{\max}/10^{2.5}$.

Figure 9: Sparse-Group Lasso experiments on climate data NCEP/NCAR Reanalysis 1 (dense data with $n = 814$ observations and $p = 73577$ features) with $\tau = 0.4$ chosen by cross-validation.

We choose the parameter $\tau$ in the set $\{0, 0.1, \ldots, 0.9, 1\}$ by splitting in half the observations, and run a training-test validation procedure. For each value of $\tau$, we require a duality gap of $10^{-8}$ on the training set and pick the best one in term of prediction accuracy on the test set. Since the prediction error degrades increasingly for $\lambda \leqslant \lambda_{\max}/10^{-2.5}$, we fix $\delta = 2.5$. We have fixed the weight $w_g = 1$ since all groups have the same size. The computational time benchmark is presented in Figure 9(c). Here also, we observe a significant gain by using a dynamic Gap Safe screening rule, which is further improved by the active warm start.

## 7. Conclusion

We have proposed a unified presentation of the Gap Safe screening rules for accelerating algorithms solving supervised learning problems under sparsity constraints. The proposed approach applies to many popular estimators that boil down to convex optimization problems where the data fitting term has a Lipschitz gradient and the regularization term is a separable sparsity enforcing function. We have shown that our methodology is more flexible than previously known safe rules as it conveniently unifies both regression and classification settings. The efficiency of the Gap Safe rules along with the new *active /strong warm start* strategies was demonstrated on multiple experiments using real high dimensional data set, suggesting that Gap Safe screening rules are always helpful to speed-up solvers targeting sparse regularization.

### Acknowledgments

This work was supported by the ANR THALAMEEG ANR-14-NEUC-0002-01, the NIH R01 MH106174, by the ERC Starting Grant SLAB ERC-YStG-676943, by the Chair Machine Learning for Big Data at Télécom ParisTech and by the Orange/Télécom ParisTech think tank Phi-TAB. We would like to thank the reviewers for their valuable comments which contributed to improve the quality of this paper.